%% file: main.tex
\documentclass{article}

\usepackage[accepted]{icml2026} % If accepted, instead use the following line for the camera-ready submission

\input{preamble}

\begin{document}

\input{01_title}
\input{02_abstract}
\input{03_introduction}
\input{04_foundations}
\input{05_models}
\input{06_training}
\input{07_benchmarks}
\input{08_evidences}
\input{09_action}

\input{10_views}
\input{11_conclusions}


\bibliography{icml2026}
\bibliographystyle{icml2026}

%%%%%%%%%%%%%%%%%%%%%%%%%%%%%%%%%%%%%%%%%%%%%%%%%%%%%%%%%%%%%%%%%%%%%%%%%%%%%%%
% APPENDIX
%%%%%%%%%%%%%%%%%%%%%%%%%%%%%%%%%%%%%%%%%%%%%%%%%%%%%%%%%%%%%%%%%%%%%%%%%%%%%%%
\input{12_appendix}

\end{document}

%% file: preamble.tex
\usepackage[utf8]{inputenc} % allow utf-8 input
\usepackage[T1]{fontenc}    % use 8-bit T1 fonts
\usepackage{xcolor} % 提供自定义颜色能力，请确保已引入
\definecolor{CVPRBlue}{RGB}{25, 88, 166}
\definecolor{CVPRPurple}{RGB}{150, 0, 80}
\usepackage[colorlinks,linkcolor=CVPRPurple,citecolor=CVPRBlue,urlcolor=black]{hyperref}
\usepackage{url}            % simple URL typesetting
\usepackage{booktabs}       % professional-quality tables
\usepackage{amsfonts}       % blackboard math symbols
\usepackage{nicefrac}       % compact symbols for 1/2, etc.
\usepackage{microtype}      % microtypography
\usepackage{multirow}
\usepackage{graphicx}      % for \resizebox
\usepackage{wrapfig}
\usepackage{enumitem}
\usepackage[table]{xcolor}
\usepackage{colortbl}
\usepackage{subcaption}
\usepackage{amsmath,amsthm,amssymb}
\usepackage{mathtools}
\usepackage[capitalize,noabbrev]{cleveref}

\usepackage{tikz}
\usetikzlibrary{mindmap,trees,shadows}
\usepackage[edges]{forest}

\usepackage[x11names, RGB]{xcolor}
\usepackage[most]{tcolorbox}

\def\algo#1{{\textbf{#1}}} % algorithm
\def\chkp#1{{\texttt{#1}}} % check point
\definecolor{academicblue}{RGB}{54, 95, 145}

\definecolor{c1}{RGB}{102,178,255} % fresh light blue
\definecolor{c2}{RGB}{255,153,153} % soft coral red
\definecolor{c3}{RGB}{255,204,102} % mellow orange
\definecolor{c4}{RGB}{153,221,153} % soft mint green
\definecolor{c5}{RGB}{204,179,255} % soft lavender purple
\definecolor{c7}{RGB}{153,221,214} % soft turquoise (更柔和)
\definecolor{c8}{RGB}{221,160,221} % plum purple
\definecolor{c9}{RGB}{255,179,207} % pastel pink

\tikzset{forked edges/.style={edge from parent forked}}

\usepackage{xparse}

\definecolor{questionyellow}{RGB}{255,204,102}

\newcommand{\iclrBadge}[2]{%
  \begingroup
  \setlength{\fboxsep}{1.3pt}%
  \colorbox{#1!18}{\bfseries\strut #2}%
  \endgroup
}

\newcommand{\iclrRuninStartNoTitle}[4]{%
  \par\addvspace{3pt}\noindent
  \begingroup
  #4%
  \iclrBadge{#1}{#2}%
  \hspace{0.55em}%
  \ignorespaces
}
\newcommand{\iclrRuninEnd}{%
  \par\endgroup
  \addvspace{3pt}%
}

\NewDocumentEnvironment{QuestionBox}{O{}m}
  {\iclrRuninStartNoTitle{c2}{QUESTION}{#2}{#1}} % questionyellow
  {\iclrRuninEnd\vspace{-0.5em}}

\NewDocumentEnvironment{TakeawayBox}{O{}m}
  {\iclrRuninStartNoTitle{c4}{TAKEAWAY}{#2}{#1}}
  {\iclrRuninEnd\vspace{-0.5em}}

\theoremstyle{plain}

\theoremstyle{definition}

\theoremstyle{remark}

\usepackage[textsize=tiny]{todonotes}

\renewcommand{\paragraph}[1]{\noindent\textbf{#1}~}

%% file: 01_title.tex
\icmltitlerunning{Position: Text Embeddings Should Capture Implicit Semantics, Not Just Surface Meaning}

\twocolumn[
  \icmltitle{Position: Text Embeddings Should Capture Implicit Semantics, \\Not Just Surface Meaning}

  % The \author macro works with any number of authors. There are two commands
  % used to separate the names and addresses of multiple authors: \And and \AND.
  %
  % Using \And between authors leaves it to LaTeX to determine where to break the
  % lines. Using \AND forces a line break at that point. So, if LaTeX puts 3 of 4
  % authors names on the first line, and the last on the second line, try using
  % \AND instead of \And before the third author name.



  \icmlsetsymbol{equal}{*}

  \begin{icmlauthorlist}
    \icmlauthor{Yiqun Sun}{NUS}
    \icmlauthor{Qiang Huang}{HITsz}
    \icmlauthor{Anthony K. H. Tung}{NUS}
    \icmlauthor{Jun Yu}{HITsz}
  \end{icmlauthorlist}

  \icmlaffiliation{NUS}{National University of Singapore}
  \icmlaffiliation{HITsz}{Harbin Institute of Technology (Shenzhen)}
  % \icmlaffiliation{NUS}{School of Computing, National University of Singapore, Singapore}
  % \icmlaffiliation{HITsz}{School of Intelligence Science and Engineering, Harbin Institute of Technology (Shenzhen)}

  \icmlcorrespondingauthor{Qiang Huang}{huangqiang@hit.edu.cn}
  
  \icmlkeywords{Machine Learning, ICML}

  \vskip 0.3in
]



\printAffiliationsAndNotice{}  % no special notice (required even if empty)

%% file: 02_abstract.tex
\begin{abstract}
\textbf{This position paper argues that text embedding research should move beyond surface meaning and embrace implicit semantics as a central modeling objective.}
%%% background
Text embeddings are a foundational component of modern NLP, underpinning a wide range of applications and driving sustained research progress.
%%% gaps between computer science and linguistics
Despite rapid progress, most embedding models remain narrowly focused on surface-level semantics, whereas linguistic theory emphasizes that much of human meaning is implicit, shaped by pragmatics, speaker intent, and sociocultural context.
%%% limitations from embedding models, datasets, and benchmarks
Current models are typically trained on datasets that lack such depth and evaluated using benchmarks that reward surface similarity. 
As a result, they struggle with tasks that require interpretive reasoning, stance recognition, or socially grounded understanding.
%%% experiments
Our pilot study makes this limitation explicit, showing that even state-of-the-art embeddings achieve only marginal improvements over simple lexical baselines on tasks probing implicit semantics.
%%% call for action
We therefore call for a paradigm shift: embedding research should prioritize linguistically grounded and diverse training data, develop benchmarks that probe deeper semantic understanding, and treat implicit meaning as a core modeling objective to better align embeddings with real-world language complexity.
%%%
The code is available at \url{http://github.com/dukesun99/Implicit-Embeddings}.
\end{abstract}

%% file: 03_introduction.tex
\section{Introduction}
\label{sect:intro}

\begin{figure}[t]
  \centering
  \includegraphics[width=0.95\columnwidth]{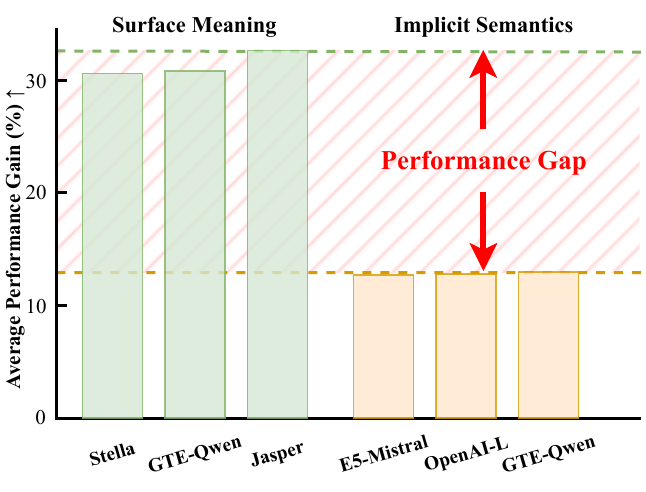}
  \caption{\textbf{Average gains over a lexical baseline (Bag-of-Tokens) for SOTA embedding models on Surface Meaning versus Implicit Semantics.} Surface performance is averaged over MTEB classification tasks~\cite{muennighoff-etal-2023-mteb}, while implicit-semantics performance is averaged over seven datasets probing pragmatic, attitudinal, and social meaning (Table~\ref{tab:benchmark-dataset-results}). The comparison illustrates a performance divide: modern embeddings show large gains on surface-semantic benchmarks but much smaller gains on tasks requiring implicit interpretation.}
  \label{fig:head} 
\end{figure}

Text embedding models map sentences, passages, or documents into dense vectors whose proximity reflects semantic similarity \cite{reimers-gurevych-2019-sentence, muennighoff-etal-2023-mteb, sun2025one}. 
%%%
They underpin much of modern NLP and Information Retrieval, serving as foundational components in Clustering \cite{grootendorst2022bertopic, huang2023new, angelov2024topic, li2026weight}, Classification \cite{muennighoff-etal-2023-mteb, enevoldsen2025mmteb}, Dense Retrieval \cite{thakur2beir, karpukhin2020dense, sun2024diversinews, huang2024diversity, tang2025uncovering}, and Retrieval-Augmented Generation (RAG) \cite{lewis2020retrieval, you2026cut, you2026knowledge, dai2026mg, sun2026don}. 
%%%
As a result, embedding models are widely deployed in a pre-trained, off-the-shelf manner and treated as general-purpose semantic interfaces for downstream decision-making.

This central role has driven rapid progress across model architectures \cite{reimers-gurevych-2019-sentence, li2024your, behnamghader2024llm2vec}, training objectives \cite{thirukovalluru2024sumcse, li-li-2024-aoe, xianmingese}, and large-scale evaluation benchmarks \cite{muennighoff-etal-2023-mteb, han2025ateb, enevoldsen2025mmteb}.
%%%
By standard benchmarks, modern embedding models appear increasingly strong, robust, and general-purpose.

\paragraph{The Overlooked Dimension: Implicit Semantics}
Despite this progress, we argue that contemporary embedding research remains narrowly focused on surface meaning, such as semantic signals derived from lexical overlap, syntactic alternation, and topical similarity, while systematically underrepresenting the implicit semantics that are fundamental to human language understanding.
%%%
Decades of linguistic research demonstrate that meaning is frequently conveyed indirectly, shaped by pragmatic inference, speaker intent, stance-taking, and sociocultural context rather than explicit propositional content \cite{huang2017introduction, ma2025pragmatics, kiesling2022stance, silverstein2003, bucholtz2005}. 
Such implicit meanings determine not only what is said, but how and why it is understood in context.

Crucially, these dimensions of meaning are not peripheral edge cases. They govern everyday interpretation in domains such as persuasion, ideology, politeness, sarcasm, safety, and social signaling. 
Yet they remain largely invisible to embedding models optimized for surface-level similarity.

\paragraph{Why Embedding Models Miss Implicit Meaning}
This limitation is structural rather than incidental.
%%%
First, dominant training corpora provide little supervision for implicit semantics. 
Most embedding models are trained on retrieval, entailment, or paraphrase datasets \cite{bajaj2016ms, kwiatkowski-etal-2019-natural}, where success is defined by lexical relevance or literal semantic equivalence rather than contextual interpretation.
%%%
Second, evaluation benchmarks overwhelmingly reward surface alignment. 
Widely used suites rarely test whether embeddings distinguish implied intent, speaker stance, or socially grounded meaning \cite{thakur2beir, muennighoff-etal-2023-mteb}.

As a result, embedding models are optimized for what is easy to measure rather than what is linguistically consequential.
Even advanced embeddings, despite high benchmark scores, are neither trained nor evaluated to capture the deeper layers of meaning that humans routinely infer.

\paragraph{A Performance Divide}
To examine this gap empirically, we conduct a pilot study spanning three tiers of implicit semantics: (1) utterance-level pragmatic inference, (2) speaker-level stance, and (3) society-level political and social meaning.
%%%
Across a suite of linguistically motivated datasets, we find that leading embedding models perform only marginally better than sparse lexical baselines when implicit understanding is required.

Figure~\ref{fig:head} summarizes this contrast. While modern embeddings achieve substantial gains over Bag-of-Tokens representations on surface-meaning benchmarks, their improvements nearly collapse on implicit semantics tasks.
%%%
This performance divide highlights a fundamental mismatch between benchmark success and interpretive competence.

\paragraph{Our Position} 
\textbf{We argue that text embedding research must move beyond surface-level semantics and treat implicit meaning as a first-class modeling objective.} 
%%%
This shift is essential for applications where semantic similarity alone is insufficient, such as stance-aware retrieval, ideological differentiation, or safety-critical filtering of content that appears benign on the surface but is implicitly harmful.

Our position is not to replace existing paradigms, but to broaden the community's understanding of what embeddings should represent.
Aligning modeling objectives with linguistic theory and real-world interpretive demands can better capture the complexity of human communication.

%% file: 04_foundations.tex
\section{Linguistic Foundations of Implicit Meaning}
\label{sect:foundations}

To clarify what we mean by \emph{implicit meaning}, we draw on linguistic theory and organize it into a three-tier framework spanning the utterance, speaker, and society levels. % utterance (pragmatics), speaker (stance-taking), and society (sociolinguistics).
We emphasize that this framework is intended as an analytical lens for organizing NLP-relevant phenomena, rather than a strict ontology or exhaustive linguistic definition of implicit meaning.
This framework highlights how meaning systematically extends beyond literal content, emerging from pragmatic inference, speaker positioning, and sociocultural context.

\begin{QuestionBox}{Implicit Semantics}
How do linguistic theories explain meaning that is conveyed but not explicitly stated?
\end{QuestionBox}

%%% Utterance Level
\paragraph{Utterance Level: Pragmatic Sources of Implicit Meaning}
Pragmatics investigates how utterances derive meaning from context, bridging the gap between literal surface semantics and a speaker's intended message \cite{grice1975logic, huang2017introduction, ma2025pragmatics}. 
Rather than focusing solely on what is explicitly said, pragmatics emphasizes what is left unsaid yet reliably understood, revealing interpretive layers that surface-level semantic analysis cannot fully capture.
%%%
This view has long influenced NLP research on tasks that require contextual reasoning and inference \cite{hovy2021importance, cambria2024pragmatics}.

A central insight of pragmatics is that meaning emerges from shared background knowledge, social norms, and situational context, which jointly constrain interpretation \cite{huang2017introduction, ma2025pragmatics}. 
%%%
Within this framework, speakers frequently rely on \emph{implicature}--meanings inferred indirectly rather than stated outright \cite{grice1975logic, potts2015presupposition, hoyle2023natural, ma2025pragmatics}. 
%%%
For instance, the sentence ``\textit{Bart managed to pass the test}'' implies that his success was unexpected, even though this is not logically entailed.

Another key mechanism is \emph{presupposition}, where utterances embed background assumptions required for comprehension~\cite{potts2015presupposition, ma2025pragmatics}. 
A statement like ``\textit{Sam quit smoking}'' presupposes that Sam previously smoked, an assumption that persists even under negation or questioning.
%%%
Together, these phenomena illustrate how meaning depends not only on explicit content but also on what listeners infer or take for granted, posing a fundamental challenge for text embeddings that aim to model such nuance.

%%% Speaker Level
\paragraph{Speaker Level: Stance and Implicit Meaning}
While pragmatics focuses on utterances in context, \emph{stance} focuses on the speaker's internal positioning: attitudes, evaluations, and degrees of alignment or commitment \cite{kiesling2022stance}. 
%%%
Stance-taking is central to implicit meaning because it conveys emotional and social orientation through subtle linguistic cues rather than explicit statements.

Linguistic research characterizes stance along three dimensions: evaluation (positive or negative appraisal), alignment (social positioning relative to others), and investment (the strength of speaker commitment) \cite{du2008stance, lempert2008poetics}. 
%%%
These dimensions are often expressed through sociolinguistic variation.
For instance, forms such as ``\textit{-in'}'' versus ``\textit{-ing}'' function not merely as dialectal variants, but as signals of informality, toughness, or solidarity \cite{kiesling2009style, trudgill1972sex}. 
%%%
Over time, such forms may become detached from specific groups and reused more broadly to index stance.
The evolution of the word ``\textit{dude},'' from a gendered term to a marker of casual camaraderie, exemplifies this process \cite{kiesling2004dude}. 

Importantly, stance is dynamic.
Quantitative studies show that speakers shift stance across discourse as they adjust intent and alignment \cite{kiesling2018reddit}. 
%%%
This introduces a relational and affective layer of meaning that complements pragmatics but remains difficult for embedding models to capture, as it depends on speaker intent rather than propositional content alone.

%%% Society Level
\paragraph{Society Level: Sociocultural Foundations of Implicit Meaning}
Beyond individual utterances and speakers, sociolinguistics examines how meaning is shaped by identity, power, and culture.
%%%
Variation in pronunciation, grammar, or vocabulary--such as dropping the ``\textit{g}'' in ``\textit{workin'},'' regional vowel shifts, or particles like ``\textit{lah}'' in Singapore English--functions as a social index, signaling class, group membership, or regional identity \cite{silverstein2003}.
%%%
These signals are culturally contingent: the same linguistic form may index friendliness in one context and stigma in another.

Language ideologies further shape implicit meaning by privileging certain varieties while marginalizing others \cite{bourdieu1991}.
%%%
Because high-status registers dominate most pretraining corpora, embedding models risk encoding and amplifying existing social hierarchies--for example, by systematically disadvantaging African-American Vernacular English relative to Standard English.
%%%
Speakers also engage in style-shifting, alternating registers, dialects, or slang to negotiate identity and social relationships \cite{bucholtz2005}. 
These shifts carry implicit meaning, signaling authority, solidarity, or deference.

Static embeddings, which average across usage, struggle to capture such rapid shifts of meaning. 
Capturing the societal dimension of language, therefore, requires sensitivity to culturally embedded cues beyond surface form.

\begin{TakeawayBox}{Linguistic Layers of Implicit Meaning}
Implicit meaning unfolds across three interconnected layers: 
(1) \textbf{pragmatics} captures what is implied but unsaid at the utterance level;  
(2) \textbf{stance-taking} reveals the speaker's evaluative and relational positioning; and  
(3) \textbf{sociolinguistics} exposes how language encodes identity, culture, and power. 
%%%
Together, these layers underscore that meaning is deeply contextual, socially embedded, and dynamically constructed, posing a significant challenge for text embeddings still anchored in surface-level representations.
\end{TakeawayBox}

%% file: 05_models.tex
\section{Text Embedding Models: Landscape and Limitations}
\label{sect:models}

\input{figures/taxonomy}

Text embedding--the task of mapping text into dense vector representations--has long been central to NLP and now underpins many state-of-the-art applications. 
This section reviews the evolution of embedding models, summarizes active research directions, and critically examines the field's current limitations.
%%%
Figure \ref{fig:embedding_taxonomy_full} provides a high-level overview of major model classes and emerging trends that shape today's embedding landscape.

\begin{QuestionBox}{Research Focus}
What is the current state of research on text embedding models?
\end{QuestionBox}

\paragraph{Early Models}
Early embedding approaches relied on static word vectors, such as Word2Vec and GloVe~\cite{NIPS2013_9aa42b31, pennington-etal-2014-glove}, pooled into sentence-level representations. 
%%%
Subsequent models, including Skip-Thought~\cite{kiros2015skip}, InferSent~\cite{conneau-etal-2017-supervised}, Sent2Vec~\cite{pagliardini-etal-2018-unsupervised}, and the Universal Sentence Encoder~\cite{cer-etal-2018-universal}, explore recurrent, transformer, or bilinear architectures to better capture sentence-level semantics. 
%%%
ELMo~\cite{peters-etal-2018-deep} marks a shift toward contextualized embeddings, dynamically encoding word meaning based on surrounding context.

\paragraph{Encoder-Only Models}
Pretrained encoder-only Transformers, such as BERT~\cite{devlin-etal-2019-bert} and RoBERTa~\cite{liu2019roberta}, further advanced sentence embedding by enabling context-aware representations via pooled token embeddings. 
These models are typically optimized using contrastive or denoising objectives \cite{zhang2025role}. 
%%%
Building on this foundation, methods such as Sentence-BERT~\cite{reimers-gurevych-2019-sentence}, SimCSE~\cite{gao-etal-2021-simcse}, TSDAE~\cite{wang2021tsdae}, and E5~\cite{wang2022text} substantially improved embedding quality through refined training strategies, better negative sampling, and architectural adjustments. 
As a result, encoder-only models remain the dominant choice in many retrieval and similarity-based applications due to their efficiency and strong benchmark performance~\cite{zhuo-etal-2023-whitenedcse, koukounas2024jina, lai2024enhancing, xiao2024pixel, li2024improving, ponwitayarat2024space, sturua2025jina, chen2024bge, mohr2024multi, gunther-etal-2023-jina, gunther2023jina}.

\paragraph{LLMs as Embedders}
More recently, LLMs have been adapted for embedding tasks using both decoder-only and encoder-decoder designs. 
Research has explored fine-tuning, prompting, and hybrid objectives to repurpose general-purpose LLMs for dense semantic representations.  
%%%
These efforts include adapting decoder-only models~\cite{muennighoff2022sgpt, cheng2025contrastive, lin-etal-2025-look}, leveraging encoder-decoder architectures~\cite{ni2022large}, or and building embedding-specific LLM variants~\cite{muennighoff2024generative, lee2024nv, lee2024gecko, behnamghader2024llm2vec, springer2024repetition, li2024bellm, man2024ullme, deng2025following, zhao2025prompt, fu2024token, yamada2025out, zhang2025cse, li2024your, zhang2025gem}. 
%%%
While such models often demonstrate strong semantic capabilities, converting generative LLMs into embedders typically requires adaptation with contrastive, retrieval, or similarity-based objectives, which may improve surface-semantic alignment without fully preserving the reasoning-relevant or implicit-semantic information available in the underlying model; their large size and high inference cost also sustain demand for lighter, encoder-based alternatives.

\paragraph{Emerging Research Directions}
Several trends characterize recent progress in text embeddings.
Instruction-following~\cite{su-etal-2023-one, peng2024answer, yoo2024hyper, weller2024followir, feng-etal-2025-dont, zhuang2025towards, romero2025beyond}, few-shot embedding~\cite{li2024making}, and ``thinking-enhanced'' representations~\cite{ji2025learning} aim to improve adaptability across tasks.
%%%
Multilingual and cross-lingual models~\cite{wang2024multilingual, zhang2024mgte, yu2024arctic, sturua2025jina, chen2024bge, mohr2024multi} extend embedding utility beyond English, while new architectural designs address efficiency, political bias, and late interaction \cite{yoo2024hyper, coelho2024dwell, sun-etal-2025-prism, khattab2020colbert, santhanam2022colbertv2, jha2024jina, kusupati2022matryoshka, xianmingese, zhuang2024starbucks}.
%%%
Interpretability has also emerged as an important concern, with growing efforts to produce embeddings that align more closely with human-understandable concepts~\cite{jha2018interpretable, senel2018semantic, subramanian2018spine, panigrahi-etal-2019-word2sense, opitz-frank-2022-sbert, simhi-markovitch-2023-interpreting, mcinerney-etal-2023-chill, Benara2024CraftingIE, o2024disentangling, huang2023bridging, cqgmbqa, zhao2026partially}. 

Another prominent direction involves distilling knowledge from LLMs into lightweight embedding models. 
This includes mining hard negatives \cite{moreira2024nv, pan2025negative}, generating synthetic training data \cite{zhang2023contrastive, wang2023improving, lee2024gecko, sato2024improving, he2025refining, gill2025advancing}, and transferring supervision from more accurate but slower teacher models \cite{tamber2025teaching, wang2025multi, ananthakrishnan2025can, thirukovalluru2024sumcse, tamber2024can}. 
%%%
While these techniques expand supervision and improve performance, they largely reinforce surface-level semantic alignment, with limited attention to implicit meaning.

\begin{TakeawayBox}{Landscape of Text Embedding Models}
Research on text embeddings spans architectures, training paradigms, multilinguality, interpretability, and efficiency. 
Yet, despite this rapid progress, the ability to capture implicit semantics--central to real-world language understanding--remains significantly underexplored. 
This gap motivates our subsequent analysis of training signals and evaluation practices, our pilot empirical study, and the research agenda outlined in the following sections.
\end{TakeawayBox}

%% file: figures/taxonomy.tex
\tikzstyle{my-box}=[
    rectangle,
    draw=gray!50,
    rounded corners,
    text opacity=1,
    minimum height=10em,
    minimum width=5em,
    inner sep=8pt, %5pt,
    inner ysep=10pt, %8pt,
    align=left,
    fill opacity=0.15,
    line width=0.5pt,
]

\tikzstyle{leaf}=[
    my-box,
    minimum height=2em,
    fill=gray!5,
    text=black,
    align=left, % only the text in leaf should be left-aligned
    font=\normalsize,
    inner xsep=5pt,
    inner ysep=8pt,
    line width=1.0pt,
]

\tikzset{forked edges/.style={edge from parent forked}}

\begin{figure*}[!t]
  \centering
  \resizebox{0.99\textwidth}{!}{
    \begin{forest}
      forked edges,
      for tree={
        grow=east,
        reversed=true,
        anchor=west,          % point taken from the middle of the left side
        parent anchor=east,   % point taken from the middle of the right side
        child anchor=west,    % (same for the child)
        base=center,
        font=\large,
        rectangle,
        draw=gray,
        rounded corners,
        text centered, % keep text center
        minimum width=8em,
        edge+={darkgray, line width=0.5mm},
        s sep=3pt,
        inner xsep=5pt,
        inner ysep=8pt,
        line width=1.0pt,
        text width=35em, % control the text width of root (gray box)
        ver/.style={rotate=90, child anchor=north, parent anchor=south, anchor=center},
      },
      where level=1{font=\normalsize,text width=10em}{},
      where level=2{font=\normalsize,text width=10em}{},
      % where level=3{font=\normalsize,text width=14em}{},
      where level=3{font=\normalsize,text width=55em}{},
      [
        {\Large \textbf{Current Research in Text Embedding}}, ver, line width=0.7mm
        [% Text Embedding Models
          {\large \textbf{Embedding Models}},
          fill=c1!60, draw=c1, line width=0.7mm
          [
            {\large \textbf{Early Models}},
            fill=c1!60, draw=c1, line width=0.7mm
            [
                GloVe \cite{pennington-etal-2014-glove}; Word2Vec \cite{NIPS2013_9aa42b31}; Skip-Thought \cite{kiros2015skip}; InferSent \cite{conneau-etal-2017-supervised}; USE \cite{cer-etal-2018-universal}; Sent2Vec \cite{pagliardini-etal-2018-unsupervised}; ELMo \cite{peters-etal-2018-deep};, 
                leaf, fill=c1!40, draw=c1
            ]
          ]
          [
            {\large \textbf{Encoder-Only Models}},
            fill=c1!60, draw=c1, line width=0.7mm
            [
                SBERT \cite{reimers-gurevych-2019-sentence}; SimCSE \cite{gao-etal-2021-simcse}; TSDAE \cite{wang2021tsdae}; WhitenedCSE \cite{zhuo-etal-2023-whitenedcse}; Angle \cite{li-li-2024-aoe}; GTR \cite{ni2022large}; E5 \cite{wang2022text}; Jina-Embedding Series \cite{gunther-etal-2023-jina, gunther2023jina, mohr2024multi, sturua2025jina}; GCSE \cite{lai2024enhancing}; Pixel Linguist \cite{xiao2024pixel}; SLERP \cite{li2024improving}; mGTE \cite{zhang2024mgte}; mE5 \cite{wang2024multilingual}; M3-Embedding \cite{chen2024bge}; MixSP \cite{ponwitayarat2024space}; EmbeddingGemma \cite{vera2025embeddinggemma};  ,
                leaf, fill=c1!40, draw=c1
            ]
          ]
          [
            {\large \textbf{Large Language Models}},
            fill=c1!60, draw=c1, line width=0.7mm
            [
                SGPT \cite{muennighoff2022sgpt}; encoder-decoder LLMs \cite{ni2022large}; UDEVER \cite{zhang2023language}; GRITLM \cite{muennighoff2024generative}; NV-Embed \cite{lee2024nv}; Gecko \cite{lee2024gecko}; LLM2Vec \cite{behnamghader2024llm2vec}; Echo \cite{springer2024repetition}; BeLLM \cite{li2024bellm}; ULLME \cite{man2024ullme}; AutoRegEmbed \cite{deng2025following}; SPT \cite{zhao2025prompt}; Token Prepending \cite{fu2024token}; PonTE \cite{yamada2025out}; CSE-SFP \cite{zhang2025cse}; Mixture-of-Experts-Based \cite{li2024your}; GenEOL \cite{thirukovalluru2024geneol}; NV-Retriever \cite{moreira2024nv}; QAEA-DR \cite{tan2025qaea}; KV-Embedding \cite{tang2026kv}; BERT-JEPA \cite{gillin2026bert}; EffiR \cite{lei2025making}; Conan-embedding-v2 \cite{li2025conan}; Causal2Vec \cite{lin2025causal2vec}; GEM \cite{zhang2025gem}; \citet{lin-etal-2025-look}; CP \cite{cheng2025contrastive}; ,
                leaf, fill=c1!40, draw=c1
            ]
          ]
          [
            {\large \textbf{Emerging Directions}},
            fill=c1!60, draw=c1, line width=0.7mm
            [
                Instruction-following \cite{su-etal-2023-one, peng2024answer, yoo2024hyper, weller2024followir, romero2025beyond, zhuang2025towards, feng-etal-2025-dont}; Few-shot examples \cite{li2024making}; Thinking-enhanced \cite{ji2025learning, zhang2025your, gui2025search, chen2025reasonembed}; Multilingual Embedding Models \cite{wang2024multilingual, zhang2024mgte, yu2024arctic, sturua2025jina, chen2024bge, mohr2024multi, janeiro2025mixture, man2025lusifer, nacar2025gate}; Multi-Modal Embedding Models \cite{koukounas2024jina, zhang2025jasperstelladistillationsota, miao2024enhancing}; 
                Positional bias \cite{coelho2024dwell}; Name bias \cite{manchanda2025name}; Political bias \cite{sun-etal-2025-prism}; Late Interaction Models \cite{khattab2020colbert, jha2024jina, santhanam2022colbertv2}; Matryoshka \cite{kusupati2022matryoshka, ayad2025compressed}; 2D Matryoshka \cite{xianmingese,zhuang2024starbucks}; Interpretability \cite{jha2018interpretable, senel2018semantic, subramanian2018spine, panigrahi-etal-2019-word2sense, opitz-frank-2022-sbert, simhi-markovitch-2023-interpreting, mcinerney-etal-2023-chill, Benara2024CraftingIE, o2024disentangling, huang2023bridging, cqgmbqa}; LLM distillation \cite{zhang2023contrastive,wang2023improving,lee2024gecko,sato2024improving,he2025refining,gill2025advancing, chen2024little, sun2025grace}; Cross-encoder teachers \cite{tamber2025teaching,wang2025multi,ananthakrishnan2025can}; API-only distill \cite{tamber2024can}; Implicit Semantics \cite{oda2025one}; Longer context \cite{wu2025sitemb}; Sticky tokens \cite{chen2025sticking}; Hard-negative mining \cite{moreira2024nv, pan2025negative}; , 
                leaf, fill=c1!40, draw=c1
            ]
          ]
        ]
        [% Training Datasets
          \large \textbf{Training Processes},
          fill=c2!60, draw=c2, line width=0.7mm
          [
            \large \textbf{Self-Supervised Learning},
            fill=c2!60, draw=c2, line width=0.7mm
            [
                SimCSE \cite{gao-etal-2021-simcse}; DenoSent \cite{wang2024denosent}; Angle-based Contrastive Learning \cite{jeong2024simple}; Dimension-wise contrastive \cite{pappadopulo2024non}; SoftCSE \cite{zhuang2024not}; SumCSE \cite{thirukovalluru2024sumcse}; GCSE \cite{lai2024enhancing}; Pixel Linguist \cite{xiao2024pixel}; ,
                leaf, fill=c2!40, draw=c2
            ]
          ]
          [
            \large \textbf{Supervised Learning},
            fill=c2!60, draw=c9, line width=0.7mm
              [
                \large \textbf{Semantic Textual Similarity},
                fill=c9!60, draw=c9, line width=0.7mm, text width=10em,
                [
                    STS17 \cite{sts17}; ,
                    leaf, fill=c9!40, draw=c9, text width=42.7em,
                ]
              ]
              [
                \large \textbf{Natural Language Inference},
                fill=c9!60, draw=c9, line width=0.7mm, text width=10em,
                [
                    SNLI \cite{snli}; MultiNLI \cite{williams2018broad}; ANLI \cite{nie2020adversarial}; QA-NLI \cite{demszky2018transforming}; ,
                    leaf, fill=c9!40, draw=c9, text width=42.7em,
                ]
              ]
              [
                \large \textbf{Information Retrieval},
                fill=c9!60, draw=c9, line width=0.7mm, text width=10em,
                [
                    MS MARCO \cite{bajaj2016ms}; NQ \cite{kwiatkowski-etal-2019-natural}; Community QA \cite{ni2022large}; HotpotQA \cite{yang2018hotpotqa}; MIRACL \cite{zhang2023miracl};,
                    leaf, fill=c9!40, draw=c9, text width=42.7em,
                ]
              ]
              [
                \large \textbf{Multi-Task Learning},
                fill=c9!60, draw=c9, line width=0.7mm, text width=10em,
                [
                    C-MTP \cite{xiao2024c}; MEDI2 \cite{muennighoff2024generative}; Jina series \cite{gunther-etal-2023-jina,gunther2023jina,sturua2025jina}; CCPairs \cite{wang2022text};,
                    leaf, fill=c9!40, draw=c9, text width=42.7em,
                ]
              ]
              % [
              %   \large \textbf{Distillation-Based Training},
              %   fill=c9!60, draw=c9, line width=0.7mm, text width=12.7em,
              %   [
              %       LLM distillation \cite{zhang2023contrastive,wang2023improving,lee2024gecko,sato2024improving,he2025refining,gill2025advancing, chen2024little}; Cross-encoder teachers \cite{tamber2025teaching,wang2025multi,ananthakrishnan2025can}; API-only distill \cite{tamber2024can};,
              %       leaf, fill=c9!40, draw=c9, text width=25em,
              %   ]
              % ]
          ]
        ]
        [% Downstream Tasks
          \large \textbf{Benchmark Suites},
          fill=c3!60, draw=c3, line width=0.7mm
          [
            \large \textbf{Semantic Textual Similarity Benchmarks},
            fill=c3!60, draw=c3, line width=0.7mm
            [
                STS12–17 \cite{sts12,sts13,sts14,sts15,sts16,sts17}; SICK-R \cite{sick-r}; STS-B \cite{huggingface:dataset:stsb_multi_mt}; MRPC \cite{dolan2005automatically}; QQP \cite{sharma2019natural}; GIS \cite{li-li-2024-aoe}; Sentence Smith \cite{li2025sentence}; Semantic-KG \cite{wei2025semantic}; MUSTS \cite{ranasinghe2025musts}; ,
                leaf, fill=c3!40, draw=c3
            ]
          ]
          [
            \large \textbf{Information Retrieval Benchmarks},
            fill=c3!60, draw=c3, line width=0.7mm
            [
                MS MARCO \cite{bajaj2016ms}; NQ \cite{kwiatkowski-etal-2019-natural}; SQuAD \cite{rajpurkar2016squad}; Mr. TyDi \cite{zhang-etal-2021-mr}; MIRACL \cite{zhang2023miracl}; BEIR \cite{thakur2beir}; BRIGHT \cite{su2024bright}; LitSearch \cite{ajith2024litsearch}; COIR \cite{li2024coir}; Scandinavian \cite{enevoldsen2024scandinavian}; BEIR-NL \cite{banar2024beir}; RusBEIR \cite{kovalev2025building}; ComLQ \cite{xu2025comlq}; ,
                leaf, fill=c3!40, draw=c3
            ]
          ]
          [
            \large \textbf{Multi-Task Benchmarks},
            fill=c3!60, draw=c3, line width=0.7mm
            [
                MTEB \cite{muennighoff-etal-2023-mteb}; C-MTEB \cite{xiao2024c}; MMTEB \cite{enevoldsen2025mmteb}; ArabicMTEB \cite{bhatia2024swan}; PL-MTEB \cite{poswiata2024pl}; FaMTEB \cite{zinvandi2025famteb}; SciRepEval \cite{singh2023scirepeval}; FinMTEB \cite{tang2025finmteb}; ChemTEB \cite{kasmaee2024chemteb}; MIEB \cite{xiao2025mieb}; ATEB \cite{han2025ateb}; Search Arena \cite{sharifymoghaddam2025chatbot}; SEA-BED \cite{ponwitayarat2025sea}; ,
                leaf, fill=c3!40, draw=c3
            ]
          ]
          [
            \large \textbf{Retrieval Augmented Generation Benchmarks},
            fill=c3!60, draw=c3, line width=0.7mm
            [
                RAG \cite{lewis2020retrieval}; RGB \cite{chen2024benchmarking}; CRUD-RAG \cite{lyu2025crud}; DomainRAG \cite{wang2024domainrag}; MultiHop-RAG \cite{tang2024multihop}; LegalBench-RAG \cite{pipitone2024legalbench}; MIRAGE-Medicine \cite{xiong2024benchmarking}; CyberMetric \cite{tihanyi2024cybermetric}; RAGBench \cite{friel2024ragbench}; RAGTruth \cite{niu2024ragtruth}; UDA \cite{hui2024uda}; BERGEN \cite{rau2024bergen}; eRAG \cite{salemi2024evaluating}; MIRAGE-Metric \cite{park2025mirage};,
                leaf, fill=c3!40, draw=c3
            ]
          ]
        ]
      ]
    \end{forest}
  }
  \caption{\textbf{A taxonomy of text embedding research:} tracing the evolution from early static models to recent trends in encoder-based architectures, LLM adaptations, multilingual embeddings, interpretability, and LLM distillation. While research has diversified, the modeling of implicit semantics remains underexplored.}
  % \caption{Taxonomy of current research in text embedding community.}
  \label{fig:embedding_taxonomy_full}
  % \vspace{-1.0em}
\end{figure*}

%% file: 06_training.tex
\section{Training Processes Fail to Capture Implicit Semantics}
\label{sect:training}

Despite architectural advances, current training signals for text embeddings overwhelmingly supervise surface-level similarity rather than implicit meaning.
%%%
This section examines the two dominant paradigms: self-supervised and supervised learning, and shows how both rely on datasets and objectives that privilege lexical overlap, paraphrastic equivalence, or relevance matching, leaving deeper pragmatic, attitudinal, and social meanings largely unmodeled.

\begin{QuestionBox}{Training Gap}
How are text embedding models trained, and why do these methods fail to capture implicit meaning?
\end{QuestionBox}

\subsection{Self-Supervised Learning}
\label{sect:training:self-supervised}

Self-supervised learning extracts training signals directly from unlabeled text using augmentation or structural cues, without requiring manual labels.
%%%
Representative approaches include SimCSE~\cite{gao-etal-2021-simcse}, which creates positive pairs via dropout noise, and denoising-based objectives such as DenoSent~\cite{wang2024denosent}. 
%%%
More recent variants explore alternative formulations, including angle-based objectives~\cite{jeong2024simple}, dimension-wise contrastive loss~\cite{pappadopulo2024non}, and similarity-weighted negative sampling~\cite{zhuang2024not}.

While self-supervised methods are attractive due to their scalability and low annotation cost, they consistently underperform supervised approaches in semantic benchmarks.
%%%
As a result, most practical embedding pipelines adopt a two-stage strategy: self-supervised pretraining followed by supervised fine-tuning~\cite{gao-etal-2021-simcse}.
Crucially, because the self-supervised signal is derived from surface perturbations of the same text, it primarily reinforces invariance to form rather than sensitivity to implicit intent or context.

\subsection{Supervised Learning}
\label{sect:training:supervised}

Supervised embedding models use contrastive objectives such as triplet, SimCSE, or angle-based losses~\cite{reimers-gurevych-2019-sentence, gao-etal-2021-simcse, li-li-2024-aoe} on pretrained language models, but rely on task-specific datasets, primarily Semantic Textual Similarity (STS), Natural Language Inference (NLI), and Information Retrieval (IR), due to the scarcity of labeled pairs in general-purpose corpora \cite{raffel2020exploring}.

\paragraph{Semantic Textual Similarity (STS)}
STS datasets provide fine-grained similarity scores and have been widely used in early sentence embedding models \cite{reimers-gurevych-2019-sentence, wang2021tsdae}. 
%%%
However, their limited scale and narrow domain coverage often lead to overfitting and weak generalization \cite{muennighoff-etal-2023-mteb}.
More importantly, STS annotations primarily reflect surface-level paraphrasing rather than deeper interpretive meaning.

\paragraph{Natural Language Inference (NLI)}
NLI datasets label sentence pairs as entailment, contradiction, or neutral and offer greater scale and diversity \cite{snli, williams2018broad, nie2020adversarial, demszky2018transforming}. 
%%%
They are widely used in modern embedding models \cite{reimers-gurevych-2019-sentence, wang2021tsdae, wang2022text, zhang2023language, li-li-2024-aoe, zhang2023language}. 
%%%
However, the semantic signal is often shallow: many entailment pairs differ only lexically or syntactically.
For instance, pairs like ``\textit{A boy is jumping on a skateboard}'' and ``\textit{The boy does a skateboarding trick}'' test literal equivalence rather than pragmatic intent \cite{snli}.

\paragraph{Information Retrieval (IR)}
Retrieval datasets such as MS MARCO~\cite{bajaj2016ms} and Natural Questions (NQ)~\cite{kwiatkowski-etal-2019-natural} dominate large-scale embedding training \cite{ni2022large, wang2022text, muennighoff2024generative, moreira2024nv, zhang2024mgte}. 
%%%
These datasets are effective for learning lexical relevance and topic matching, but they reward literal overlap between queries and documents.
As a result, embeddings trained on IR objectives struggle to encode implicit cues such as stance, ideology, sarcasm, or social framing.

\paragraph{Multi-Task Learning}
To improve robustness, recent models such as mGTE \cite{zhang2024mgte} and Jina \cite{gunther-etal-2023-jina, gunther2023jina, sturua2025jina} adopt multi-task training that combines STS, NLI, IR, QA, and related datasets \cite{xiao2024c, muennighoff2024generative}.
%%%
While this broadens task and domain coverage, supervision remains largely surface-oriented, with few examples probing pragmatic inference, speaker stance, or sociocultural meaning.

\begin{TakeawayBox}{Training Data Emphasize Surface Semantics}
Across both self-supervised and supervised paradigms, current embedding training pipelines are anchored in datasets that prioritize surface-level similarity. Although multi-task learning increases diversity, it does not fundamentally change what models are optimized to learn.
As a result, core components of implicit meaning, i.e., pragmatics, stance, and social context, remain weakly represented or entirely missing from training objectives.
\end{TakeawayBox}

%% file: 07_benchmarks.tex
\section{Benchmarks Do Not Evaluate Implicit Semantics}
\label{sect:evaluation}

Despite the rapid expansion of large-scale benchmark suites--from semantic similarity and retrieval to multi-task generalization--most evaluations remain focused on surface-level semantics.
%%%
This section surveys widely used STS datasets, retrieval, multi-task, and RAG benchmarks, highlighting a persistent gap: the limited evaluation of implicit, contextual, and socially situated meaning.

\begin{QuestionBox}{Evaluation Gap}
How are text embedding models evaluated, and why do existing benchmarks fall short in capturing implicit meaning?
\end{QuestionBox}

\paragraph{STS Benchmarks}
evaluate how well model-predicted similarities align with human judgments, typically using correlation-based metrics. 
%%%
Popular datasets include STS12--17~\cite{sts12, sts13, sts14, sts15, sts16, sts17}, STS-B~\cite{huggingface:dataset:stsb_multi_mt}, and SICK-R~\cite{sick-r}, along with related paraphrase classification tasks such as MRPC~\cite{dolan2005automatically}, QQP~\cite{sharma2019natural}, and GIS~\cite{li-li-2024-aoe}. 
%%%
While STS tasks can probe deeper meaning, they are largely constrained to lexical and syntactic variation, rarely testing pragmatic, attitudinal, or cultural interpretation.  

\paragraph{IR Benchmarks}
IR benchmarks evaluate how effectively embeddings retrieve relevant documents using ranking metrics such as MRR, nDCG, and recall@$k$~\cite{wang2013theoretical, thakur2beir}.
%%%
Datasets like MS MARCO \cite{bajaj2016ms}, Natural Questions \cite{kwiatkowski-etal-2019-natural}, SQuAD \cite{rajpurkar2016squad}, Mr. TyDi \cite{zhang-etal-2021-mr}, and MIRACL~\cite{zhang2023miracl} are commonly used, with BEIR~\cite{thakur2beir} aggregating many such tasks across diverse retrieval settings. 
More recent benchmarks extend coverage to new domains and languages \cite{su2024bright, ajith2024litsearch, li2024coir, enevoldsen2024scandinavian, banar2024beir, kovalev2025building}.
%%%
Despite this breadth, IR benchmarks focus on surface relevance and seldom evaluate alignment with implicit criteria such as stance or ideology.
Retrieval based on speaker stance or ideological framing remains largely unexplored.

\paragraph{Multi-Task Benchmarks}
MTEB~\cite{muennighoff-etal-2023-mteb} seldom test whether retrieved documents align with implicit criteria such as stance, ideological framing, or communicative intent \cite{xiao2024c, enevoldsen2025mmteb, bhatia2024swan, poswiata2024pl, zinvandi2025famteb, singh2023scirepeval, tang2025finmteb, kasmaee2024chemteb, xiao2025mieb}.
%%%
More challenging variants introduce reasoning or instruction-following tasks~\cite{han2025ateb}, and crowdsourced platforms like MTEB Arena\footnote{\url{https://huggingface.co/spaces/mteb/arena}} and Search Arena\footnote{\url{https://blog.lmarena.ai/blog/2025/search-arena/}} provide user-driven comparisons \cite{sharifymoghaddam2025chatbot}. 
%%%
Despite their scale and flexibility, these benchmarks rely largely on surface-level metrics and datasets.
Only a small subset probes beyond lexical meaning, so strong MTEB performance often reflects surface alignment rather than sensitivity to implicit meaning.

\paragraph{RAG Benchmarks}
RAG benchmarks assess how well embeddings support retrieval for downstream generation.
%%%
Existing benchmarks cover multilingual, domain-specific, and multi-hop scenarios \cite{chen2024benchmarking, lyu2025crud, wang2024domainrag, tang2024multihop, pipitone2024legalbench, xiong2024benchmarking, tihanyi2024cybermetric}, and include tools for evaluating retrieval quality and hallucination \cite{friel2024ragbench, niu2024ragtruth, hui2024uda, rau2024bergen, salemi2024evaluating, park2025mirage}. 
%%%
While these setups introduce more complex pipelines, the underlying retrieval objectives remain largely factual.
As a result, the underlying semantic evaluations resemble IR tasks, offering limited insight into how well embeddings reflect implicit intent, stance, or social meaning.

\begin{TakeawayBox}{Current Evaluation Focuses Mostly on Surface Semantics}
Current benchmarks provide extensive task and domain coverage, but overwhelmingly emphasize surface-level similarity and relevance.
They rarely assess a model's ability to capture implicit meaning, such as pragmatics, stance, or social context, leaving a critical gap in how we evaluate semantic understanding.
\end{TakeawayBox}

%% file: 08_evidences.tex
\section{Empirical Evidences}
\label{sect:evidences}

\input{tables/results_aspect_level}
To ground our argument empirically and motivate future research, we conduct a pilot study examining whether state-of-the-art embedding models capture implicit meaning across three linguistic tiers: utterance, speaker, and society.

\paragraph{Experimental Setup}
We evaluate embeddings on seven datasets spanning three levels of implicit semantics:
(1) Utterance level: Pragmatics Understanding Benchmark (PUB), including Implicature (\textbf{P-IMP}), Presupposition (\textbf{P-PRE}), and Reference \& Deixis (\textbf{P-R\&D})~\cite{sravanthi2024pub, louis2020d, zheng2021grice, liu2022testing, chakrabarty2022flute, jeretic2020natural, parrish2021nope}; 
(2) Speaker level: \textbf{P-Stance} dataset~\cite{li2021p} for stance detection; and 
(3) Society level: the datasets of Implicit Hate Speech (\textbf{IHS})~\cite{elsherief2021latent}, Social Bias Inference Corpus (\textbf{SBIC})~\cite{sap2020social}, and Political Bias (\textbf{Pol. Bias})~\cite{baly2020we}. 
%%%
To provide a structured comparison, we report mean performance across subtasks within each semantic tier.

Since these datasets were not originally designed for embedding evaluation, we reformulate them into classification, pairwise classification (following the MTEB protocol~\cite{muennighoff-etal-2023-mteb}), and zero-shot similarity settings, where models select labels based on embedding similarity.
%%%
We evaluate four representative model families: 
(1) encoder-only models, (2) LLM-based embeddings, (3) multimodal encoders, and (4) proprietary embeddings (OpenAI). 
As baselines, we include a Bag-of-Tokens lexical model \cite{harris1954distributional, cqgmbqa} and a random predictor (Random) as a hardness baseline.
Full implementation details are provided in Appendix~\ref{app:impl}.

\paragraph{Results and Analysis}
As depicted in Table~\ref{tab:benchmark-dataset-results}, the results reveal a consistent and striking pattern.
%%%
Encoder-only models typically perform only marginally better than the Bag-of-Tokens baseline and, in some cases, approach random performance.
%%%
In contrast, LLM-based models and proprietary embeddings achieve stronger overall results. 
Notably, although OpenAI embeddings rank lower on standard benchmarks such as MTEB, they perform comparatively well on implicit semantics tasks, suggesting a disconnect between benchmark success and deeper semantic competence.

Performance also varies substantially across semantic tiers.
Linq-Mistral excels at utterance-level pragmatic tasks, OpenAI-Large performs best on speaker- and society-level datasets, and E5-Mistral shows particular strength in political bias detection.
These differences suggest that current models specialize unevenly across dimensions of implicit meaning, rather than learning a unified representation.

These positive results also clarify the nature of the limitation. 
Current embedding models are not uniformly poor on all implicit semantics tasks; rather, they perform better when the relevant implicit meaning is strongly associated with local lexical, syntactic, discourse, or label-specific cues. 
This is especially plausible for relatively conventionalized pragmatic phenomena, where large-scale pretraining and instruction tuning may help models recover recurring surface patterns correlated with the intended label. 
Thus, the issue is not complete failure, but uneven generalization: current embeddings appear to capture some shallow or partially lexicalized forms of implicit meaning, while remaining less robust on cases that require richer contextual grounding, speaker modeling, or socially situated interpretation.

These results support our central claim: \textbf{state-of-the-art embedding models remain limited in their ability to capture implicit semantics.}
%%%
High MTEB scores do not translate into robustness on tasks involving pragmatic inference, stance recognition, or social meaning. 
The shrinking gains over Bag-of-Tokens, especially on the hardest pragmatic tasks, underscore a fundamental gap between how embeddings are evaluated and how meaning is actually constructed in language.
%%%
Detailed results are provided in Appendix \ref{app:results}.

%% file: tables/results_aspect_level.tex
\begin{table*}[t]
\centering
\small
\renewcommand{\arraystretch}{1.3}
\caption{\textbf{Average accuracy (\%) of representative embedding models} on seven datasets spanning three tiers of implicit semantics: utterance-level pragmatics, speaker-level stance, and societal-level social meaning. Results reveal substantial variation across semantic tiers and demonstrate that strong performance on standard benchmarks does not translate to robust modeling of implicit meaning.}
\label{tab:benchmark-dataset-results}
\resizebox{1.0\textwidth}{!}{
\begin{tabular}{lcccccccc}
    \toprule
    \multirow{2.5}{*}{\textbf{Model}} & \multicolumn{3}{c}{\textbf{Utterance Level}} & \multicolumn{1}{c}{\textbf{Speaker Level}} & \multicolumn{3}{c}{\textbf{Society Level}} & \multirow{2.5}{*}{\textbf{\shortstack{Average \\ Accuracy $\uparrow$}}} \\
    \cmidrule(lr){2-4} \cmidrule(lr){5-5} \cmidrule(lr){6-8}
    & \textbf{P-IMP} & \textbf{P-PRE} & \textbf{P-R\&D} & \textbf{P-Stance} & \textbf{IHS} & \textbf{SBIC} & \textbf{Pol. Bias} & \\
    \midrule
    \algo{Random} & 48.5 & 54.1 & 38.8 & 51.3 & 27.5 & 59.2 & 34.5 & 44.8 \\
    \algo{Bag-of-Tokens} \cite{cqgmbqa} & 56.5 & 75.3 & 48.2 & 73.4 & 59.6 & 80.7 & 41.6 & 62.2 \\
    \midrule
    \rowcolor[RGB]{255, 244, 230}
    \multicolumn{9}{c}{\textit{\textbf{Encoder-only Models}}} \\
    \midrule
    \algo{S-BERT} \cite{reimers-gurevych-2019-sentence} & 61.7 & 72.8 & 55.7 & 72.9 & 60.8 & 81.8 & 47.9 & 64.8 \\
    \algo{GIST-Small} \cite{solatorio2024gistembed} & 65.8 & 76.1 & 58.8 & 76.0 & 61.8 & 81.6 & 49.1 & 67.0 \\
    \algo{BGE-Base} \cite{bge_embedding} & 65.0 & 75.6 & 57.3 & 74.4 & 62.9 & 82.1 & 52.1 & 67.1 \\
    \algo{Angle} \cite{li-li-2024-aoe} & 69.4 & 78.8 & 57.2 & 76.4 & 59.5 & 83.7 & 50.4 & 67.9 \\
    \algo{BGE-Large} \cite{bge_embedding} & 68.3 & 75.5 & 58.1 & 76.0 & 63.5 & 83.4 & 51.5 & 68.0 \\
    \algo{MXBAI-Large} \cite{emb2024mxbai} & 69.8 & 78.2 & 59.4 & 75.6 & 60.1 & 83.6 & 50.5 & 68.2 \\
    \algo{GIST-Large} \cite{solatorio2024gistembed} & 68.8 & 76.6 & 62.9 & 76.7 & 64.2 & 83.4 & 52.5 & 69.3 \\
    \algo{Stella} \cite{zhang2025jasperstelladistillationsota} & 72.1 & 81.5 & 59.6 & 76.5 & 60.4 & 84.0 & 54.4 & 69.8 \\
    \midrule
    \rowcolor[HTML]{E7F3FF} %\rowcolor[RGB]{255, 244, 230}
    \multicolumn{9}{c}{\textit{\textbf{LLM-based Embeddings}}} \\
    \midrule
    \algo{Linq-Mistral} \cite{LinqAIResearch2024} & \textbf{80.3} & \textbf{87.7} & \textbf{70.4} & 75.8 & 61.4 & 82.0 & 56.8 & 73.5 \\
    \algo{E5-Mistral} \cite{wang2023improving, wang2022text} & 78.1 & 81.8 & 63.4 & 81.1 & 63.9 & 84.8 & \textbf{71.5} & 74.9 \\
    \algo{GTE-Qwen} \cite{li2023towards} & 73.4 & 87.3 & 68.1 & 80.9 & 65.6 & 84.5 & 66.8 & \textbf{75.2} \\
    \midrule
    \rowcolor[HTML]{D9EAD3} %\rowcolor[RGB]{255, 244, 230}
    \multicolumn{9}{c}{\textit{\textbf{Multimodal Encoders}}} \\
    \midrule
    \algo{Jasper} \cite{zhang2025jasperstelladistillationsota} & 73.3 & 80.1 & 63.0 & 80.1 & 65.7 & 84.2 & 63.9 & 72.9 \\ 
    \midrule
    \rowcolor[HTML]{FFF2CC} %\rowcolor[RGB]{255, 244, 230}
    \multicolumn{9}{c}{\textit{\textbf{Proprietary Embeddings}}} \\
    \midrule
    \algo{OpenAI-Small} \cite{neelakantan2022text} & 71.3 & 78.1 & 64.3 & 80.0 & 66.2 & 83.9 & 56.6 & 71.5 \\
    \algo{OpenAI-Large} \cite{neelakantan2022text} & 76.0 & 80.2 & 66.4 & \textbf{83.7} & \textbf{67.1} & \textbf{85.4} & 66.3 & 75.0 \\
    \bottomrule
\end{tabular}}
\end{table*}

%% file: 09_action.tex
\section{Towards Embeddings that Capture Implicit Meaning}
\label{sect:future}

Building on the empirical findings in Section \ref{sect:evidences}, we outline three directions for advancing text embeddings beyond surface-level semantics: 
(1) linguistically and culturally diverse training data,
(2) benchmarks that directly evaluate pragmatic, attitudinal, and social understanding, and
(3) treating implicit semantics as a core modeling objective. 
%%%
Together, these directions realign embedding research with the realities of human language interpretation.

\subsection{Curating More Diverse Training Data}
\label{sect:future:data}

Training data ultimately determines what embedding models can learn.
When supervision emphasizes surface-level signals, representations inevitably mirror surface meaning.
Capturing implicit semantics, therefore, requires expanding beyond narrow, convenience-driven datasets toward linguistically richer and culturally diverse sources. 

Importantly, by more diverse training data, we do not simply mean scaling raw web text.
The key need is more diverse supervision for embedding learning.
Unlike autoregressive LMs, embedding models are usually trained with pairwise, contrastive, retrieval-style, or classification-oriented signals.
For implicit semantics, such supervision should include cases where surface meaning is similar but implied meaning differs, such as contrastive examples involving implicature, presupposition, stance, indirectness, sarcasm, social framing, dialectal variation, or ideology.

Recent progress in LLM-based data generation offers a practical path forward.
Prior work shows that LLMs can synthesize effective supervision for embedding training, often by enforcing semantic invariance or contrast~\cite{wang2023improving, chen2024little, lippmann2025zero, truong2025learning}. 
Future efforts should target implicit phenomena such as implicature, presupposition, stance, and indirect evaluation, rather than treating them as incidental variation.
Such data can be curated from dialogue, indirect answers, stance-rich discourse, and socially situated language, or synthesized and relabeled using LLMs or stronger cross-encoder teachers guided by linguistic theory.

Linguistic theory provides a rich foundation for this shift. 
Decades of research have formalized typologies of implicit meaning across pragmatic and sociocultural dimensions. 
Aligning synthetic and curated data with these frameworks can help embedding models internalize forms of meaning that are largely absent from existing training corpora.

\subsection{Designing Benchmarks for Implicit Meaning}
\label{sect:future:benchmarks}

Benchmarks shape research priorities by defining what success looks like.
%%%
Existing suites, most notably MTEB, have played an important role in standardizing evaluation, but they overwhelmingly emphasize surface similarity.
Their open and leaderboard-driven nature has also led to data leakage and score inflation, weakening their ability to measure generalization \cite{chung2025maintaining, sancheti2025less}.

Empirical studies show that strong MTEB results do not reliably translate to downstream robustness and may even correlate negatively with performance on certain tasks. 
This trend reflects a broader shift toward optimizing benchmark artifacts rather than transferable semantic understanding.

To address this gap, new benchmarks should explicitly target implicit meaning. 
This includes tasks that require inference from indirect cues, recognition of speaker stance, sensitivity to sociolinguistic variation, and interpretation of socially situated language. 
Without such benchmarks, progress on implicit semantics will remain invisible and undervalued.

\subsection{Framing Implicit Semantics as a Modeling Goal}
\label{sect:future:goal}

A deeper issue is that implicit meaning is rarely treated as a first-class objective in embedding research.
While work on LLMs increasingly targets pragmatic reasoning, social understanding, and discourse-level interpretation~\cite{li2020molweni, li2023diplomat, kazemi2023boardgameqa, sravanthi2024pub, yue2024large, curry2024classist, sun2024diversinews, cqgmbqa, ma2025pragmatics}, embedding models continue to be optimized for objectives that reward superficial alignment.

This misalignment pushes models to optimize what is easy to measure rather than what is meaningful to understand, reinforcing shallow representations in the absence of explicit objectives for implicit semantics.
Reframing implicit semantics as a core modeling goal can better align embeddings with how humans interpret language in context.

Concretely, this goal can be operationalized through objectives that distinguish texts with similar explicit content but different implied meanings, multi-task supervision over pragmatics, stance, and social meaning, or distillation from models that explicitly reason over implicit interpretations.
The goal is not to make embeddings preserve all information in the original text.
Rather, it is to preserve task-relevant implicit signals that matter for downstream applications such as stance-aware retrieval, socially grounded classification, recommendation, clustering, filtering, and reranking.

Instruction-following retrieval is a particularly relevant step in this direction because it conditions embeddings on user intent and task framing rather than only surface similarity.
Recent work on instruction-following embeddings and retrieval models~\cite{su-etal-2023-one, peng2024answer, weller2024followir, feng-etal-2025-dont, zhuang2025towards} shows that embedding research is already moving toward richer task-conditioned representations.

We view this line of work as a promising bridge between standard semantic embeddings and implicit-semantics-aware representations.
Nevertheless, following retrieval instructions does not by itself guarantee sensitivity to broader implicit phenomena such as stance, presupposition, sociocultural meaning, or social indexicality.

%% file: 10_views.tex
\section{Alternative Views}
\label{sect:views}

It is reasonable to argue that surface-level semantics suffice for many practical applications, such as search, recommendation, or clustering. 
In these settings, modeling deeper meaning may introduce unnecessary complexity and additional training or evaluation costs without clear returns.
%%%
Another perspective holds that implicit, pragmatic, and socially grounded meaning is better handled by LLMs designed for contextual reasoning, while embeddings should prioritize efficiency and general-purpose utility.
From this perspective, expanding the embedding scope risks blurring their role.

We acknowledge these positions.
Our argument is not that embeddings should replace decoder-only LMs for complex reasoning, nor that all implicit understanding must be solved purely in vector space.
Rather, embeddings often serve as first-stage representations in practical systems, where they determine which examples are retrieved, clustered, recommended, filtered, or passed to downstream models.
In such settings, if important signals such as stance, intent, or social framing are not preserved in the embedding space, later components may never receive the relevant evidence.

Thus, as embeddings increasingly serve as semantic interfaces for downstream decision-making, their limitations in capturing implicit meaning become harder to ignore, especially in applications where stance, intent, or social interpretation are central.

%% file: 11_conclusions.tex
\section{Conclusions}
\label{sect:conclusions}

Despite rapid progress, contemporary text embedding models are predominantly optimized for surface-level semantics, exhibiting a fundamental limitation in capturing the implicit meanings that are essential to nuanced human communication. 
%%%
Grounded in a three-tier linguistic framework: spanning utterance-level pragmatics, speaker-level stance, and society-level sociocultural context, our empirical analysis reveals that even state-of-the-art models show only marginal gains over simple lexical baselines on tasks probing these deeper layers of interpretation.
These findings expose a structural mismatch between training, evaluation, and real-world language use. 
%%%
To bridge this gap, we advocate for semantically richer data, benchmarks that explicitly test implicit understanding, and modeling objectives that treat implicit semantics as a first-class goal. 
Aligning embedding research with these dimensions of real-world language use is crucial for developing robust, context-aware models that power the next generation of language-aware applications.

\section*{Acknowledgements}
Qiang Huang was supported by the New Generation Artificial Intelligence-National Science and Technology Major Project (2025ZD0123302) and the National Natural Science Foundation of China (NSFC) under Grant No. U25B6003. 
%%%
Jun Yu was supported by the NSFC under Grant Nos. 62125201 and U24B20174.
%%%
Yiqun Sun and Anthony T. H. Tung were supported by the Ministry of Education, Singapore, under its MOE AcRF TIER 1 Grant (T1 251RES2517) and the National Research Foundation, Singapore under its AI Singapore Programme (AISG Award No: AISG3-RP-2022-029).
%%%
Any opinions, findings, and conclusions or recommendations expressed in this material are those of the author(s) and do not reflect the views of the Ministry of Education, Singapore, or the National Research Foundation, Singapore.

%% file: 12_appendix.tex
\newpage
\appendix
\onecolumn

\section{Implementation Details}
\label{app:impl}

\subsection{Model Checkpoints} 
\label{app:impl:checkpoints}

We use the model checkpoints listed in Table~\ref{tab:model-checkpoint-mapping} for all experiments reported in Section~\ref{sect:evidences}. 
%%%
To ensure a fair and reproducible comparison, we adopt the \textbf{official checkpoints} used by the MTEB benchmark and evaluate every open-source model under the \textbf{default configurations} provided by the Sentence Transformers library~\cite{reimers-gurevych-2019-sentence}.
Importantly, we do not apply any additional parameter tuning, task-specific calibration, or prompt engineering; this design choice isolates the intrinsic capability of each embedding model under standard deployment settings.

\paragraph{Proprietary Embeddings}
For OpenAI's proprietary embedding models, we obtain representations using OpenAI’s official client library\footnote{\url{https://platform.openai.com/docs/api-reference/embeddings}} and follow the recommended embedding API usage.
This ensures that the results reflect the intended inference behavior of these models rather than custom wrappers or ad hoc post-processing.

\paragraph{Complementary Baselines}
To contextualize performance, we include two complementary baselines:
\begin{itemize}[nolistsep,left=10pt]
  \item \textbf{Random:} 
  For each dataset, we implement a random predictor by sampling labels according to the empirical label distribution. 
  This provides a hardness reference that accounts for class imbalance and prevents overstating performance on skewed datasets.
  
  \item \textbf{Bag-of-Tokens \cite{harris1954distributional, cqgmbqa}:} 
  As a lexical-feature baseline, we include a Bag-of-Tokens representation using the \chkp{google-bert/bert-base-uncased} tokenizer.
  This baseline serves as a strong surface-matching reference point, helping quantify how much modern embeddings improve beyond lexical overlap.
\end{itemize}

Together, these implementation choices are designed to make comparisons consistent across model families, while keeping the evaluation protocol aligned with widely used community standards.

\input{tables/checkpoints}

\subsection{Implicit Semantics Tasks} 
\label{app:impl:tasks}

Our evaluation targets \textbf{implicit semantics} across diverse task formulations and data sources. 
Because the underlying datasets differ in structure and annotation format, we organize the evaluation into three settings: \textbf{classification}, \textbf{pair classification}, and \textbf{zero-shot classification}. 
Each setting includes the following datasets:
\begin{itemize}[nolistsep,left=10pt]
  \item \textbf{Classification:} 
  From the \textbf{Pragmatics Understanding Benchmark (PUB)}, we include \textit{Task 1 (Direct/Indirect Classification)}, \textit{Task 2 (Response Classification without Implied Meaning)}, \textit{Task 3 (with Implied Meaning)}, \textit{Task 6 (Understanding Sarcasm)}, \textit{Task 10 (Implicature NLI)}, \textit{Task 11 (Presupposition NLI)}, \textit{Task 12 (Presupposition over QA)}, and \textit{Task 13 (Deictic QA)}. 
  %%%
  We also evaluate all three subsets of the \textbf{P-Stance} dataset--\textit{Trump}, \textit{Biden}, and \textit{Bernie}--for stance classification. 
  %%%
  For the \textbf{Implicit Hate Speech (IHS)} dataset, we include \textit{detection}, \textit{categorization}, and \textit{target identification} tasks. 
  %%%
  For the \textbf{Social Bias Inference Corpus (SBIC)}, we evaluate five binary classification tasks: \textit{whoTarget} (whether the target is a group), \textit{intentYN} (intent to offend), \textit{sexYN} (presence of sexual content), \textit{offensiveYN} (offensiveness), and \textit{hasBiasedImplication} (biased implications). 
  %%%
  Lastly, we include the \textbf{Political Bias (Pol. Bias)} classification dataset.
  
  \item \textbf{Pair Classification:} We adapt \textit{Task 5 (Agreement Detection)} from \textbf{PUB}.
  
  \item \textbf{Zero-shot Classification:} We include \textit{Task 4 (Implicature Recovery)}, \textit{Task 7 (Figurative Language Understanding--No Hint)}, \textit{Task 8 (with Positive Hint)}, \textit{Task 9 (with Contrastive Hint)}, and \textit{Task 14 (Reference via Metonymy)} from \textbf{PUB}. 
\end{itemize}

\input{tables/pub_tasks}

\subsection{Evaluation Protocols} 
\label{app:impl:eval}

For \textbf{classification} and \textbf{pair classification} tasks, we follow the standard evaluation protocol used in MTEB~\cite{muennighoff-etal-2023-mteb}. 
%%%
In this setting, embeddings are computed for inputs (and, when applicable, for candidate labels or verbalized label descriptions), and predictions are made using the corresponding embedding-based classification procedure consistent with MTEB's implementation.

For \textbf{zero-shot classification}, we follow the embedding-based approach\footnote{\url{https://platform.openai.com/docs/guides/embeddings\#use-cases}} described in OpenAI's documented use case: the input query (or question) and its associated text are embedded jointly, while each candidate answer option is embedded separately. 
The model selects the option with the highest similarity to the input embedding. This protocol provides a standardized way to test ``label selection by semantic similarity,'' allowing comparison across both open-source and proprietary embeddings.

\section{Additional Results}
\label{app:results}

The complete results, including the accuracy (\%) for individual task, are presented in Tables~\ref{tab:pub-benchmark-results} and~\ref{tab:other-benchmark-results}. 
The values reported in Table~\ref{tab:benchmark-dataset-results} are computed by averaging across tasks within each dataset.

\paragraph{Widespread Variance Across Models} 
A central observation is that performance varies substantially across both models and tasks, often in ways that are not predicted by surface-level benchmark rankings.
%%%
For example, many models achieve near-perfect accuracy on \textit{Task 1 (Direct/Indirect Classification)} from PUB, suggesting that some pragmatic cues can be captured by surface correlates. 
However, several widely used embedding models, such as \textbf{GIST-Small}, \textbf{S-BERT}, and \textbf{BGE-Base}, perform only marginally better than \textbf{Bag-of-Tokens} on more challenging tasks, and in some cases fall below it.

This pattern becomes even more pronounced on certain implicature-oriented tasks.
%%%
On \textit{Task 10 (Implicature NLI)}, multiple models, including OpenAI's proprietary models and the LLM-based models like \textbf{E5-Mistral} and \textbf{Jasper}, underperform the \textbf{Bag-of-Tokens} baseline.
%%%
This indicates that higher capacity or modern training recipes do not automatically translate into reliable pragmatic competence, and that lexical heuristics can sometimes outperform dense representations when the latter fail to encode the relevant inferential structure.
%%%
Overall, these results reinforce a key theme of the paper: \textbf{Success on conventional surface-level benchmarks does not reliably transfer to tasks requiring deeper interpretive understanding.}

\input{tables/other_tasks}

\paragraph{Strengths of Large and Multimodal Models} 
While variance is widespread, the results also reveal a consistent advantage for large-scale and multimodal embedding models.
%%%
\textbf{Jasper}, for example, ranks among the top across a wide range of tasks, with particularly strong performance on socially grounded datasets such as IHS and SBIC.
Similarly, large-scale models like \textbf{E5-Mistral} and \textbf{OpenAI-Large} perform strongly across multiple domains, including social bias classification and pragmatic reasoning tasks.

These outcomes suggest that scale and broader pretraining signals can improve robustness for complex semantic phenomena. 
%%%
However, the advantage is not uniform: even high-performing models exhibit sharp drops on specific pragmatic tasks, indicating that capacity helps but does not fully resolve the underlying challenge.

\paragraph{Persistent Challenges in Implicature and Reference Tasks} 
Despite improvements from scale and multimodality, certain pragmatic phenomena remain consistently difficult across all evaluated models.
%%%
In particular, \textit{Task 4 (Implicature Recovery)} remains difficult across all models, with scores rarely exceeding 50\%. 
Even top-tier models like \textbf{GTE-Qwen} and \textbf{OpenAI-Large} achieve only modest gains over \textbf{Bag-of-Tokens}. 

This trend points to a deeper limitation of current training pipelines. 
Many dominant training signals--whether self-supervised, NLI-based, or retrieval-based--do not systematically require recovering unstated intent or bridging pragmatic gaps. 
As a result, even powerful models may rely on shallow correlates rather than encoding the inferential structure needed for implicature and reference resolution.

\paragraph{Implications for Benchmark and Model Design} 
Taken together, the appendix results expose persistent blind spots in current embedding models, especially for tasks involving implicature, figurative language, presupposition, reference, and socially grounded inference. 
More importantly, they clarify that these weaknesses are not isolated edge cases: they reflect a systematic mismatch between (i) what embeddings are trained to optimize and (ii) what is required to represent implicit meaning.

Addressing these gaps will likely require both linguistically grounded supervision that targets implicit semantics and benchmarks that directly measure interpretive competence beyond surface similarity. 
In this sense, the appendix results provide a detailed empirical foundation for the agenda proposed in the paper.

%% file: tables/checkpoints.tex
\begin{table*}[h]
\centering
\small
\renewcommand{\arraystretch}{1.2}
\caption{\textbf{List of evaluated embedding models and the corresponding checkpoints used in our experiments.}} 
\label{tab:model-checkpoint-mapping}
\begin{tabular}{lll}
    \toprule
    \rowcolor[HTML]{FFF2CC}
    \textbf{Model} & \textbf{Model Size} & \textbf{Checkpoint} \\
    \midrule
    \algo{S-BERT} \cite{reimers-gurevych-2019-sentence} & 22.7M & \chkp{sentence-transformers\textbackslash
    all-MiniLM-L6-v2} \\
    \algo{GIST-Small} \cite{solatorio2024gistembed} & 33.4M & \chkp{avsolatorio\textbackslash GIST-small-Embedding-v0} \\
    \algo{BGE-Base} \cite{bge_embedding} & 109M & \chkp{BAAI\textbackslash bge-base-en-v1.5} \\
    \algo{Angle} \cite{li-li-2024-aoe} & 335M & \chkp{WhereIsAI\textbackslash UAE-Large-V1} \\
    \algo{BGE-Large} \cite{bge_embedding} & 335M & \chkp{BAAI\textbackslash bge-large-en-v1.5} \\
    \algo{MXBAI-Large} \cite{emb2024mxbai} & 335M & \chkp{mixedbread-ai\textbackslash mxbai-embed-large-v1} \\
    \algo{GIST-Large} \cite{solatorio2024gistembed} & 335M & \chkp{avsolatorio\textbackslash GIST-large-Embedding-v0} \\
    \algo{Stella} \cite{zhang2025jasperstelladistillationsota} & 435M & \chkp{NovaSearch\textbackslash stella\_en\_400M\_v5}  \\
    \algo{Linq-Mistral} \cite{LinqAIResearch2024} & 7.11B & \chkp{Linq-AI-Research\textbackslash Linq-Embed-Mistral} \\
    \algo{E5-Mistral} \cite{wang2023improving, wang2022text} & 7.11B & \chkp{intfloat\textbackslash e5-mistral-7b-instruct} \\
    \algo{GTE-Qwen} \cite{li2023towards} & 7.61B & \chkp{Alibaba-NLP\textbackslash gte-Qwen2-7B-instruct} \\
    \algo{Jasper} \cite{zhang2025jasperstelladistillationsota} & 1.99B & \chkp{NovaSearch\textbackslash jasper\_en\_vision\_language\_v1} \\
    \algo{OpenAI-Small} & N.A. & \chkp{text-embedding-3-small} \\
    \algo{OpenAI-Large} & N.A. & \chkp{text-embedding-3-large} \\
    \bottomrule
\end{tabular}
\end{table*}

%% file: tables/pub_tasks.tex
\begin{table*}[!t]
\centering
\small
\renewcommand{\arraystretch}{1.2}
\caption{\textbf{Accuracy (\%) of embedding models on the Pragmatics Understanding Benchmark (PUB) tasks.} Each column corresponds to one of the 14 PUB tasks (T1--T14), covering diverse pragmatic phenomena.}
\label{tab:pub-benchmark-results}
\resizebox{\textwidth}{!}{
\begin{tabular}{lcccccccccccccc}
    \toprule
    \multirow{4}{*}{\textbf{Model}} & \multicolumn{14}{c}{\textbf{Pragmatics Understanding Benchmark (PUB)}} \\
    \cmidrule(lr){2-15}
    & \multicolumn{10}{c}{\textbf{Implicature}} & \multicolumn{2}{c}{\textbf{Presupposition}} & \multicolumn{2}{c}{\textbf{Ref. \& Deixis}} \\
    \cmidrule(lr){2-11} \cmidrule(lr){12-13} \cmidrule(lr){14-15}
    & \textbf{T1} & \textbf{T2} & \textbf{T3} & \textbf{T4} & \textbf{T5} & \textbf{T6} & \textbf{T7} & \textbf{T8} & \textbf{T9} & \textbf{T10} & \textbf{T11} & \textbf{T12} & \textbf{T13} & \textbf{T14} \\
    \midrule
    \algo{Random} & 48.0 & 49.2 & 52.9 & 24.9 & 50.0 & 51.0 & 49.2 & 47.1 & 49.7 & 62.9 & 36.4 & 71.7 & 56.0 & 21.7 \\
    \algo{Bag-of-Tokens} \cite{cqgmbqa} & 81.2 & 69.4 & 78.2 & 29.3 & 51.0 & 12.5 & 50.4 & 74.7 & 32.7 & 86.0 & 66.9 & 83.6 & 64.5 & 31.9 \\
    \midrule
    \rowcolor[RGB]{255, 244, 230}
    \multicolumn{15}{c}{\textit{\textbf{Encoder-only Models}}} \\
    \midrule
    \algo{S-BERT} \cite{reimers-gurevych-2019-sentence} & 82.0 & 72.4 & 83.1 & 35.4 & 53.5 & 34.0 & 59.9 & 78.6 & 39.0 & 79.0 & 60.0 & \textbf{85.6} & 68.5 & 42.9 \\
    \algo{GIST-Small} \cite{solatorio2024gistembed} & 74.8 & 74.2 & 83.3 & 44.5 & 56.1 & 40.2 & 67.4 & 86.8 & 52.4 & 77.9 & 66.9 & 85.2 & 68.5 & 49.0 \\
    \algo{BGE-Base} \cite{bge_embedding} & 83.0 & 71.5 & 79.6 & 33.2 & 56.0 & 42.8 & 69.8 & 80.9 & 55.6 & 77.6 & 66.1 & 85.2 & 68.0 & 46.6 \\
    \algo{Angle} \cite{li-li-2024-aoe} & 97.2 & 77.3 & 81.2 & 41.5 & 58.1 & 32.8 & 75.1 & 82.8 & 60.7 & 87.6 & 74.2 & 83.4 & 66.0 & 48.4 \\
    \algo{BGE-Large} \cite{bge_embedding} & 87.0 & 76.8 & 79.8 & 43.0 & 57.7 & 41.8 & 75.3 & 84.0 & 59.6 & 77.9 & 65.8 & 85.2 & 68.0 & 48.1 \\
    \algo{MXBAI-Large} \cite{emb2024mxbai} & 97.2 & 78.4 & 81.0 & 41.2 & 58.5 & 34.0 & 75.5 & 83.3 & 61.2 & 87.4 & 73.6 & 82.8 & 67.5 & 51.2 \\
    \algo{GIST-Large} \cite{solatorio2024gistembed} & 79.6 & 76.8 & 83.1 & 46.9 & 57.9 & 37.8 & 78.0 & 88.5 & 61.5 & 78.3 & 68.1 & 85.2 & 71.5 & 54.3 \\
    \algo{Stella} \cite{zhang2025jasperstelladistillationsota} & 95.8 & 79.8 & 81.9 & 51.1 & 60.0 & 35.5 & 76.2 & 87.7 & 60.8 & 91.7 & 83.6 & 79.3 & 66.0 & 53.2 \\
    \midrule
    \rowcolor[HTML]{E7F3FF} %\rowcolor[RGB]{255, 244, 230}
    \multicolumn{15}{c}{\textit{\textbf{LLM-based Embeddings}}} \\
    \midrule
    \algo{Linq-Mistral} \cite{LinqAIResearch2024} & 99.6 & \textbf{89.3} & \textbf{88.6} & 47.5 & \textbf{70.0} & 67.0 & \textbf{88.6} & \textbf{94.5} & 61.4 & \textbf{96.2} & \textbf{91.4} & 84.0 & 74.0 & \textbf{66.7} \\
    \algo{E5-Mistral} \cite{wang2023improving, wang2022text} & 97.2 & 87.2 & 85.8 & 43.8 & 69.5 & \textbf{69.0} & 87.7 & 92.7 & \textbf{62.9} & 85.0 & 78.3 & 85.2 & 68.0 & 58.9 \\
    \algo{GTE-Qwen} \cite{li2023towards} & \textbf{100.0} & 87.7 & 87.5 & 43.6 & 61.8 & 63.0 & 69.0 & 82.7 & 48.5 & 89.8 & 89.4 & 85.2 & \textbf{78.0} & 58.2 \\
    \midrule
    \rowcolor[HTML]{D9EAD3} %\rowcolor[RGB]{255, 244, 230}
    \multicolumn{15}{c}{\textit{\textbf{Multimodal Encoders}}} \\
    \midrule
    \algo{Jasper} \cite{zhang2025jasperstelladistillationsota} & 97.8 & 84.2 & 85.4 & 50.6 & 61.6 & 49.8 & 78.0 & 88.4 & 55.4 & 81.9 & 75.0 & 85.2 & 70.5 & 55.6 \\
    \midrule
    \rowcolor[HTML]{FFF2CC} %\rowcolor[RGB]{255, 244, 230}
    \multicolumn{15}{c}{\textit{\textbf{Proprietary Embeddings}}} \\
    \midrule
    \algo{OpenAI-Small} & 98.8 & 79.4 & 84.9 & \textbf{56.0} & 56.7 & 35.5 & 79.3 & 89.9 & 55.0 & 78.1 & 71.1 & 85.2 & 73.0 & 55.6 \\
    \algo{OpenAI-Large} & 99.6 & 87.7 & 87.7 & 50.8 & 61.9 & 55.5 & 83.9 & 91.8 & 58.4 & 83.1 & 75.3 & 85.2 & 73.5 & 59.3 \\
    \bottomrule
\end{tabular}}
\end{table*}

%% file: tables/other_tasks.tex
\begin{table*}[!t]
\centering
\small
\renewcommand{\arraystretch}{1.2}
\caption{\textbf{Accuracy (\%) of embedding models on additional implicit meaning benchmarks.} \textbf{P-Stance} includes stance detection for Trump, Biden, and Bernie. \textbf{Implicit Hate Speech (IHS)} comprises detection (Det.), categorization (Cat.), and target identification (Tar.) tasks. The \textbf{Social Bias Inference Corpus (SBIC)} consists of target identification (Tar.), intent (Int.), sexism (Sex.), offensiveness (Off.), and bias detection (Bias) tasks. \textbf{Political Bias (Pol. Bias)} denotes political ideology classification.}
\label{tab:other-benchmark-results}
\resizebox{\textwidth}{!}{
\begin{tabular}{lcccccccccccc}
    \toprule
    \multirow{2.5}{*}{\textbf{Model}} & \multicolumn{3}{c}{\textbf{P-Stance}} & \multicolumn{3}{c}{\textbf{IHS}} & \multicolumn{5}{c}{\textbf{SBIC}} & \multirow{2.5}{*}{\textbf{Pol. Bias}} \\
    \cmidrule(lr){2-4}
    \cmidrule(lr){5-7}
    \cmidrule(lr){8-12}
     & \textbf{Trump} & \textbf{Biden} & \textbf{Bernie} & \textbf{Det.} & \textbf{Cat.} & \textbf{Tar.} & \textbf{Tar.} & \textbf{Int.} & \textbf{Sex.} & \textbf{Off.} & \textbf{Bias} \\
    \midrule
    \algo{Random} & 51.7 & 50.9 & 51.2 & 52.5 & 16.6 & 13.4 & 48.7 & 58.2 & 76.6 & 62.0 & 50.5 & 34.5 \\
    \algo{Bag-of-Tokens} \cite{cqgmbqa} & 74.6 & 75.4 & 70.1 & 74.5 & 55.0 & 49.3 & 77.7 & 76.1 & 92.0 & 79.6 & 78.1 & 41.6 \\
    \midrule
    \rowcolor[RGB]{255, 244, 230}
    \multicolumn{13}{c}{\textit{\textbf{Encoder-only Models}}} \\
    \midrule
    \algo{S-BERT} \cite{reimers-gurevych-2019-sentence} & 72.1 & 77.4 & 69.3 & 73.2 & 58.0 & 51.2 & 78.8 & 78.0 & 91.9 & 81.2 & 79.1 & 47.9 \\
    \algo{GIST-Small} \cite{solatorio2024gistembed} & 76.6 & 78.4 & 72.9 & 74.0 & 59.5 & 51.9 & 77.9 & 77.7 & 93.1 & 81.3 & 78.0 & 49.1 \\
    \algo{BGE-Base} \cite{bge_embedding} & 74.5 & 78.8 & 69.9 & 74.2 & 60.7 & 53.9 & 78.7 & 78.5 & 92.6 & 82.0 & 78.6 & 52.1 \\
    \algo{Angle} \cite{li-li-2024-aoe} & 77.2 & 79.9 & 72.1 & 75.9 & 56.0 & 46.6 & 80.4 & 80.4 & 93.3 & 83.7 & 80.5 & 50.4 \\
    \algo{BGE-Large} \cite{bge_embedding} & 75.8 & 80.0 & 72.1 & 75.1 & 61.4 & 53.9 & 80.3 & 79.9 & 93.3 & 83.0 & 80.6 & 51.5 \\
    \algo{MXBAI-Large} \cite{emb2024mxbai} & 76.4 & 78.9 & 71.5 & 76.4 & 56.5 & 47.3 & 80.4 & 80.3 & 93.1 & 83.6 & 80.4 & 50.5 \\
    \algo{GIST-Large} \cite{solatorio2024gistembed} & 76.8 & 80.1 & 73.2 & 75.0 & 63.1 & 54.4 & 80.5 & 79.3 & 93.3 & 82.9 & 80.9 & 52.5 \\
    \algo{Stella} \cite{zhang2025jasperstelladistillationsota} & 79.2 & 80.0 & 70.2 & 76.9 & 56.7 & 47.6 & 81.2 & 80.9 & 93.4 & 83.2 & 81.5 & 54.4 \\
    \midrule
    \rowcolor[HTML]{E7F3FF} %\rowcolor[RGB]{255, 244, 230}
    \multicolumn{13}{c}{\textit{\textbf{LLM-based Embeddings}}} \\
    \midrule
    \algo{Linq-Mistral} \cite{LinqAIResearch2024} & 79.4 & 78.8 & 69.3 & 75.1 & 57.8 & 51.2 & 79.7 & 79.1 & 89.8 & 81.9 & 79.8 & 56.8 \\
    \algo{E5-Mistral} \cite{wang2023improving, wang2022text} & 84.8 & 82.3 & 76.1 & 79.2 & 61.7 & 50.9 & 82.0 & \textbf{82.5} & 93.3 & 83.9 & 82.1 & \textbf{71.5} \\
    \algo{GTE-Qwen} \cite{li2023towards} & 83.8 & 82.0 & 76.9 & 79.1 & 63.2 & 54.5 & 82.1 & 80.6 & 94.0 & 83.7 & 82.0 & 66.8 \\
    \midrule
    \rowcolor[HTML]{D9EAD3} %\rowcolor[RGB]{255, 244, 230}
    \multicolumn{13}{c}{\textit{\textbf{Multimodal Encoders}}} \\
    \midrule
    \algo{Jasper} \cite{zhang2025jasperstelladistillationsota} & 81.6 & 82.6 & 76.2 & 78.4 & 64.6 & 54.1 & 81.4 & 81.2 & 93.9 & 83.1 & 81.5 & 63.9 \\
    \midrule
    \rowcolor[HTML]{FFF2CC} %\rowcolor[RGB]{255, 244, 230}
    \multicolumn{13}{c}{\textit{\textbf{Proprietary Embeddings}}} \\
    \midrule
    \algo{OpenAI-Small} & 82.4 & 81.1 & 76.7 & 78.5 & 64.7 & \textbf{55.2} & 80.7 & 81.1 & 93.5 & 83.3 & 80.8 & 56.6 \\
    \algo{OpenAI-Large} & \textbf{87.5} & \textbf{83.8} & \textbf{79.7} & \textbf{80.2} & \textbf{67.3} & 53.7 & \textbf{82.9} & 82.3 & \textbf{94.2} & \textbf{84.7} & \textbf{83.1} & 66.3 \\
    \bottomrule
\end{tabular}}
\end{table*}

%% file: icml2026.bib
@inproceedings{muennighoff-etal-2023-mteb,
    title = "{MTEB}: Massive Text Embedding Benchmark",
    author = "Muennighoff, Niklas  and
      Tazi, Nouamane  and
      Magne, Loic  and
      Reimers, Nils",
    booktitle = "Proceedings of the 17th Conference of the European Chapter of the Association for Computational Linguistics (EACL)",
    year = "2023",
    pages = "2014--2037",
}

@article{ma2025pragmatics,
  title={Pragmatics in the era of large language models: A survey on datasets, evaluation, opportunities and challenges},
  author={Ma, Bolei and Li, Yuting and Zhou, Wei and Gong, Ziwei and Liu, Yang Janet and Jasinskaja, Katja and Friedrich, Annemarie and Hirschberg, Julia and Kreuter, Frauke and Plank, Barbara},
  journal={arXiv preprint arXiv:2502.12378},
  year={2025}
}

@incollection{grice1975logic,
  title={Logic and conversation},
  author={Grice, Herbert P},
  booktitle={Speech Acts},
  pages={41--58},
  year={1975},
  publisher={Brill}
}

@incollection{huang2017introduction,
  author       = {Huang, Yan},
  title        = {Introduction: What is Pragmatics?},
  booktitle    = {The Oxford Handbook of Pragmatics},
  editor       = {Huang, Yan},
  series       = {Oxford Handbooks},
  year         = {2017},
  publisher    = {Oxford University Press},
  url          = {https://doi.org/10.1093/oxfordhb/9780199697960.013.33},
  urldate      = {2025-05-07}
}

@incollection{cambria2024pragmatics,
  title={Pragmatics Processing},
  author={Cambria, Erik},
  booktitle={Understanding Natural Language Understanding},
  pages={229--338},
  year={2024}
}

@inproceedings{hovy2021importance,
  title={The importance of modeling social factors of language: Theory and practice},
  author={Hovy, Dirk and Yang, Diyi},
  booktitle={Proceedings of the 2021 Conference of the North American Chapter of the Association for Computational Linguistics: Human Language Technologies (NAACL-HLT)},
  pages={588--602},
  year={2021}
}

@inproceedings{hoyle2023natural,
  title={Natural Language Decompositions of Implicit Content Enable Better Text Representations},
  author={Hoyle, Alexander and Sarkar, Rupak and Goel, Pranav and Resnik, Philip},
  booktitle={Proceedings of the 2023 Conference on Empirical Methods in Natural Language Processing},
  pages={13188--13214},
  year={2023}
}

@article{potts2015presupposition,
  title={Presupposition and implicature},
  author={Potts, Christopher},
  journal={The handbook of contemporary semantic theory},
  pages={168--202},
  year={2015},
  publisher={Wiley Online Library}
}

@article{kiesling2022stance,
  title={Stance and stancetaking},
  author={Kiesling, Scott F},
  journal={Annual Review of Linguistics},
  volume={8},
  number={1},
  pages={409--426},
  year={2022}
}

@incollection{du2008stance,
  title={The stance triangle},
  author={Du Bois, John W},
  booktitle={Stancetaking in Discourse: Subjectivity, Evaluation, Interaction},
  pages={139--182},
  year={2008}
}

@article{lempert2008poetics,
 title={The poetics of stance: Text-metricality, epistemicity, interaction},
  author={Lempert, Michael},
  journal={Language in Society},
  volume={37},
  number={4},
  pages={569--592},
  year={2008}
}

@article{kiesling2004dude,
  title={Dude},
  author={Kiesling, Scott F},
  journal={American Speech},
  volume={79},
  number={3},
  pages={281--305},
  year={2004}
}

@article{trudgill1972sex,
  title={Sex, covert prestige and linguistic change in the urban British English of Norwich},
  author={Trudgill, Peter},
  journal={Language in Society},
  volume={1},
  number={2},
  pages={179--195},
  year={1972}
}

@article{kiesling2018reddit,
  title={Interactional stancetaking in online forums},
  author={Kiesling, Scott F and Pavalanathan, Umashanthi and Fitzpatrick, Jim and Han, Xiaochuang and Eisenstein, Jacob},
  journal={Computational Linguistics},
  volume={44},
  number={4},
  pages={683--718},
  year={2018}
}

@incollection{kiesling2009style,
  author = {Scott F. Kiesling},
  title = {Style as stance: Stance as the explanation for patterns of sociolinguistic variation},
  booktitle = {Stance: Sociolinguistic Perspectives},
  pages = {171--194},
  year = {2009}

}

@article{bourdieu1991,
  title={Language and Symbolic Power},
  author={Bourdieu, Pierre},
  journal={Polity},
  year={1991}
}

@article{silverstein2003,
  title={Indexical Order and the Dialectics of Sociolinguistic Life},
  author={Silverstein, Michael},
  journal={Language \& Communication},
  volume={23},
  number={3-4},
  pages={193--229},
  year={2003},
}

@article{bucholtz2005,
  title={Identity and interaction: A sociocultural linguistic approach},
  author={Bucholtz, Mary and Hall, Kira},
  journal={Discourse Studies},
  volume={7},
  number={4-5},
  pages={585--614},
  year={2005}
}

@inproceedings{kiros2015skip,
  title={Skip-thought vectors},
  author={Kiros, Ryan and Zhu, Yukun and Salakhutdinov, Ruslan and Zemel, Richard S and Torralba, Antonio and Urtasun, Raquel and Fidler, Sanja},
  booktitle={Proceedings of the 29th International Conference on Neural Information Processing Systems (NIPS)},
  pages={3294--3302},
  year={2015}
}

@inproceedings{conneau-etal-2017-supervised,
    title = "Supervised Learning of Universal Sentence Representations from Natural Language Inference Data",
    author = {Conneau, Alexis  and
      Kiela, Douwe  and
      Schwenk, Holger  and
      Barrault, Lo{\"i}c  and
      Bordes, Antoine},
    booktitle = "Proceedings of the 2017 Conference on Empirical Methods in Natural Language Processing (EMNLP)",
    pages={670--680},
    year = "2017",
}

@inproceedings{cer-etal-2018-universal,
    title = "Universal Sentence Encoder for {E}nglish",
    author = "Cer, Daniel  and
      Yang, Yinfei  and
      Kong, Sheng-yi  and
      Hua, Nan  and
      Limtiaco, Nicole  and
      St. John, Rhomni  and
      Constant, Noah  and
      Guajardo-Cespedes, Mario  and
      Yuan, Steve  and
      Tar, Chris  and
      Strope, Brian  and
      Kurzweil, Ray",
    booktitle = "Proceedings of the 2018 Conference on Empirical Methods in Natural Language Processing: System Demonstrations (EMNLP)",
    pages={169--174},    
    year = "2018",
}

@inproceedings{pennington-etal-2014-glove,
  title = "{G}lo{V}e: {G}lobal {V}ectors for {W}ord {R}epresentation",
  author = "Pennington, Jeffrey  and
      Socher, Richard  and
      Manning, Christopher",
  booktitle = "Proceedings of the 2014 Conference on Empirical Methods in Natural Language Processing ({EMNLP})",
  year = "2014",
  pages = "1532--1543",
}

@inproceedings{NIPS2013_9aa42b31,
  title={Distributed representations of words and phrases and their compositionality},
  author={Mikolov, Tomas and Sutskever, Ilya and Chen, Kai and Corrado, Greg and Dean, Jeffrey},
  booktitle={Proceedings of the 26th International Conference on Neural Information Processing Systems (NIPS)},
  pages={3111--3119},
  year={2013}
}

@article{muennighoff2022sgpt,
  title={Sgpt: Gpt sentence embeddings for semantic search},
  author={Muennighoff, Niklas},
  journal={arXiv preprint arXiv:2202.08904},
  year={2022}
}

@inproceedings{gao-etal-2021-simcse,
    title = "{S}im{CSE}: Simple Contrastive Learning of Sentence Embeddings",
    author = "Gao, Tianyu  and
      Yao, Xingcheng  and
      Chen, Danqi",
    booktitle = "Proceedings of the 2021 Conference on Empirical Methods in Natural Language Processing (EMNLP)",
    year = "2021",
    pages = "6894--6910",
}

@inproceedings{zhuo-etal-2023-whitenedcse,
    title = "{W}hitened{CSE}: Whitening-based Contrastive Learning of Sentence Embeddings",
    author = "Zhuo, Wenjie  and
      Sun, Yifan  and
      Wang, Xiaohan  and
      Zhu, Linchao  and
      Yang, Yi",
    booktitle = "Proceedings of the 61st Annual Meeting of the Association for Computational Linguistics (ACL)",
    year = "2023",
    pages = "12135--12148",
}

@inproceedings{reimers-gurevych-2019-sentence,
    title = "Sentence-{BERT}: Sentence Embeddings using {S}iamese {BERT}-Networks",
    author = "Reimers, Nils  and
      Gurevych, Iryna",
    booktitle = "Proceedings of the 2019 Conference on Empirical Methods in Natural Language Processing and the 9th International Joint Conference on Natural Language Processing (EMNLP-IJCNLP)",
    year = "2019",
    pages = "3982--3992",
}

@inproceedings{li-li-2024-aoe,
    title = "{A}o{E}: Angle-optimized Embeddings for Semantic Textual Similarity",
    author = "Li, Xianming  and
      Li, Jing",
    booktitle = "Proceedings of the 62nd Annual Meeting of the Association for Computational Linguistics (ACL)",
    year = "2024",
    pages = "1825--1839",
}

@inproceedings{devlin-etal-2019-bert,
    title = "{BERT}: Pre-training of Deep Bidirectional Transformers for Language Understanding",
    author = "Devlin, Jacob  and
      Chang, Ming-Wei  and
      Lee, Kenton  and
      Toutanova, Kristina",
    booktitle = "Proceedings of the 2019 Conference of the North {A}merican Chapter of the Association for Computational Linguistics: Human Language Technologies (NAACL-HLT)",
    year = "2019",
    pages = "4171--4186",
}

@inproceedings{wang2021tsdae,
  title={TSDAE: Using Transformer-based Sequential Denoising Auto-Encoderfor Unsupervised Sentence Embedding Learning},
  author={Wang, Kexin and Reimers, Nils and Gurevych, Iryna},
  booktitle={Findings of the Association for Computational Linguistics: EMNLP 2021},
  pages={671--688},
  year={2021}
}

@inproceedings{pagliardini-etal-2018-unsupervised,
    title = "Unsupervised Learning of Sentence Embeddings Using Compositional n-Gram Features",
    author = "Pagliardini, Matteo  and
      Gupta, Prakhar  and
      Jaggi, Martin",
    booktitle = "Proceedings of the 2018 Conference of the North {A}merican Chapter of the Association for Computational Linguistics: Human Language Technologies (NAACL-HLT)",
    pages = "528--540",
    year = "2018",
}

@inproceedings{ni2022large,
  title={Large Dual Encoders Are Generalizable Retrievers},
  author={Ni, Jianmo and Qu, Chen and Lu, Jing and Dai, Zhuyun and Abrego, Gustavo Hernandez and Ma, Ji and Zhao, Vincent and Luan, Yi and Hall, Keith and Chang, Ming-Wei and others},
  booktitle={Proceedings of the 2022 Conference on Empirical Methods in Natural Language Processing (EMNLP)},
  pages={9844--9855},
  year={2022}
}

@article{bajaj2016ms,
  title={{MS MARCO}: A {H}uman {G}enerated {MA}chine {R}eading {CO}mprehension {D}ataset},
  author={Payal Bajaj and Daniel Campos and Nick Craswell and Li Deng and Jianfeng Gao and Xiaodong Liu and Rangan Majumder and Andrew McNamara and Bhaskar Mitra and Tri Nguyen and Mir Rosenberg and Xia Song and Alina Stoica and Saurabh Tiwary and Tong Wang},
  journal={arXiv preprint arXiv:1611.09268},
  year={2016}
}

@article{kwiatkowski-etal-2019-natural,
  title={Natural questions: a benchmark for question answering research},
  author={Kwiatkowski, Tom and Palomaki, Jennimaria and Redfield, Olivia and Collins, Michael and Parikh, Ankur and Alberti, Chris and Epstein, Danielle and Polosukhin, Illia and Devlin, Jacob and Lee, Kenton and others},
  journal={Transactions of the Association for Computational Linguistics (TACL)},
  volume={7},
  pages={453--466},
  year={2019},
}

@article{wang2022text,
  title={Text embeddings by weakly-supervised contrastive pre-training},
  author={Wang, Liang and Yang, Nan and Huang, Xiaolong and Jiao, Binxing and Yang, Linjun and Jiang, Daxin and Majumder, Rangan and Wei, Furu},
  journal={arXiv preprint arXiv:2212.03533},
  year={2022}
}

@article{wang2023improving,
  title={Improving text embeddings with large language models},
  author={Wang, Liang and Yang, Nan and Huang, Xiaolong and Yang, Linjun and Majumder, Rangan and Wei, Furu},
  journal={arXiv preprint arXiv:2401.00368},
  year={2023}
}

@article{wang2024multilingual,
  title={Multilingual e5 text embeddings: A technical report},
  author={Wang, Liang and Yang, Nan and Huang, Xiaolong and Yang, Linjun and Majumder, Rangan and Wei, Furu},
  journal={arXiv preprint arXiv:2402.05672},
  year={2024}
}

@article{behnamghader2024llm2vec,
  title={Llm2vec: Large language models are secretly powerful text encoders},
  author={BehnamGhader, Parishad and Adlakha, Vaibhav and Mosbach, Marius and Bahdanau, Dzmitry and Chapados, Nicolas and Reddy, Siva},
  journal={arXiv preprint arXiv:2404.05961},
  year={2024}
}

@article{lyu2025crud,
  title={Crud-rag: A comprehensive chinese benchmark for retrieval-augmented generation of large language models},
  author={Lyu, Yuanjie and Li, Zhiyu and Niu, Simin and Xiong, Feiyu and Tang, Bo and Wang, Wenjin and Wu, Hao and Liu, Huanyong and Xu, Tong and Chen, Enhong},
  journal={ACM Transactions on Information Systems},
  volume={43},
  number={2},
  pages={1--32},
  year={2025},
}

@article{lee2024nv,
  title={Nv-embed: Improved techniques for training llms as generalist embedding models},
  author={Lee, Chankyu and Roy, Rajarshi and Xu, Mengyao and Raiman, Jonathan and Shoeybi, Mohammad and Catanzaro, Bryan and Ping, Wei},
  journal={arXiv preprint arXiv:2405.17428},
  year={2024}
}

@article{lee2024gecko,
  title={Gecko: Versatile text embeddings distilled from large language models},
  author={Lee, Jinhyuk and Dai, Zhuyun and Ren, Xiaoqi and Chen, Blair and Cer, Daniel and Cole, Jeremy R and Hui, Kai and Boratko, Michael and Kapadia, Rajvi and Ding, Wen and others},
  journal={arXiv preprint arXiv:2403.20327},
  year={2024}
}

@article{enevoldsen2024scandinavian,
  title={The scandinavian embedding benchmarks: Comprehensive assessment of multilingual and monolingual text embedding},
  author={Enevoldsen, Kenneth and Kardos, M{\'a}rton and Muennighoff, Niklas and Nielbo, Kristoffer L},
  journal={Advances in Neural Information Processing Systems (NeurIPS)},
  volume={37},
  pages={40336--40358},
  year={2024}
}

@inproceedings{xiao2024c,
  title={C-pack: Packed resources for general chinese embeddings},
  author={Xiao, Shitao and Liu, Zheng and Zhang, Peitian and Muennighoff, Niklas and Lian, Defu and Nie, Jian-Yun},
  booktitle={Proceedings of the 47th international ACM SIGIR Conference on Research and Development in Information Retrieval (SIGIR)},
  pages={641--649},
  year={2024}
}

@inproceedings{singh2023scirepeval,
  title={SciRepEval: A Multi-Format Benchmark for Scientific Document Representations},
  author={Singh, Amanpreet and D'Arcy, Mike and Cohan, Arman and Downey, Doug and Feldman, Sergey},
  booktitle={Proceedings of the 2023 Conference on Empirical Methods in Natural Language Processing (EMNLP)},
  pages={5548--5566},
  year={2023}
}

@article{springer2024repetition,
  title={Repetition improves language model embeddings},
  author={Springer, Jacob Mitchell and Kotha, Suhas and Fried, Daniel and Neubig, Graham and Raghunathan, Aditi},
  journal={arXiv preprint arXiv:2402.15449},
  year={2024}
}

@article{zhang2024mgte,
  title={mgte: Generalized long-context text representation and reranking models for multilingual text retrieval},
  author={Zhang, Xin and Zhang, Yanzhao and Long, Dingkun and Xie, Wen and Dai, Ziqi and Tang, Jialong and Lin, Huan and Yang, Baosong and Xie, Pengjun and Huang, Fei and others},
  journal={arXiv preprint arXiv:2407.19669},
  year={2024}
}

@article{su2024bright,
  title={Bright: A realistic and challenging benchmark for reasoning-intensive retrieval},
  author={Su, Hongjin and Yen, Howard and Xia, Mengzhou and Shi, Weijia and Muennighoff, Niklas and Wang, Han-yu and Liu, Haisu and Shi, Quan and Siegel, Zachary S and Tang, Michael and others},
  journal={arXiv preprint arXiv:2407.12883},
  year={2024}
}

@article{li2024making,
  title={Making text embedders few-shot learners},
  author={Li, Chaofan and Qin, MingHao and Xiao, Shitao and Chen, Jianlyu and Luo, Kun and Shao, Yingxia and Lian, Defu and Liu, Zheng},
  journal={arXiv preprint arXiv:2409.15700},
  year={2024}
}

@article{koukounas2024jina,
  title={Jina clip: Your clip model is also your text retriever},
  author={Koukounas, Andreas and Mastrapas, Georgios and G{\"u}nther, Michael and Wang, Bo and Martens, Scott and Mohr, Isabelle and Sturua, Saba and Akram, Mohammad Kalim and Mart{\'\i}nez, Joan Fontanals and Ognawala, Saahil and others},
  journal={arXiv preprint arXiv:2405.20204},
  year={2024}
}

@article{li2024coir,
  title={Coir: A comprehensive benchmark for code information retrieval models},
  author={Li, Xiangyang and Dong, Kuicai and Lee, Yi Quan and Xia, Wei and Zhang, Hao and Dai, Xinyi and Wang, Yasheng and Tang, Ruiming},
  journal={arXiv preprint arXiv:2407.02883},
  year={2024}
}

@inproceedings{su-etal-2023-one,
    title = "One Embedder, Any Task: Instruction-Finetuned Text Embeddings",
    author = "Su, Hongjin  and
      Shi, Weijia  and
      Kasai, Jungo  and
      Wang, Yizhong  and
      Hu, Yushi  and
      Ostendorf, Mari  and
      Yih, Wen-tau  and
      Smith, Noah A.  and
      Zettlemoyer, Luke  and
      Yu, Tao",
    booktitle = "Findings of the Association for Computational Linguistics: ACL 2023",
    month = jul,
    year = "2023",
    pages = "1102--1121",
}

@inproceedings{zhang2023contrastive,
  title={Contrastive Learning of Sentence Embeddings from Scratch},
  author={Zhang, Junlei and Lan, Zhenzhong and He, Junxian},
  booktitle={Proceedings of the 2023 Conference on Empirical Methods in Natural Language Processing (EMNLP)},
  pages={3916--3932},
  year={2023}
}

@inproceedings{ajith2024litsearch,
  title={LitSearch: A Retrieval Benchmark for Scientific Literature Search},
  author={Ajith, Anirudh and Xia, Mengzhou and Chevalier, Alexis and Goyal, Tanya and Chen, Danqi and Gao, Tianyu},
  booktitle={Proceedings of the 2024 Conference on Empirical Methods in Natural Language Processing (EMNLP)},
  pages={15068--15083},
  year={2024}
}

@inproceedings{miao2024enhancing,
  title={Enhancing Cross-lingual Sentence Embedding for Low-resource Languages with Word Alignment},
  author={Miao, Zhongtao and Wu, Qiyu and Zhao, Kaiyan and Wu, Zilong and Tsuruoka, Yoshimasa},
  booktitle={Findings of the Association for Computational Linguistics: NAACL 2024},
  pages={3225--3236},
  year={2024}
}

@article{enevoldsen2025mmteb,
  title={Mmteb: Massive multilingual text embedding benchmark},
  author={Enevoldsen, Kenneth and Chung, Isaac and Kerboua, Imene and Kardos, M{\'a}rton and Mathur, Ashwin and Stap, David and Gala, Jay and Siblini, Wissam and Krzemi{\'n}ski, Dominik and Winata, Genta Indra and others},
  journal={arXiv preprint arXiv:2502.13595},
  year={2025}
}

@inproceedings{li2024bellm,
  title={{BeLLM}: Backward Dependency Enhanced Large Language Model for Sentence Embeddings},
  author={Li, Xianming and Li, Jing},
  booktitle={Proceedings of the 2024 Conference of the North American Chapter of the Association for Computational Linguistics: Human Language Technologies (NAACL-HLT)},
  pages={792--804},
  year={2024}
}

@inproceedings{peng2024answer,
  title={Answer is All You Need: Instruction-following Text Embedding via Answering the Question},
  author={Peng, Letian and Zhang, Yuwei and Wang, Zilong and Srinivasa, Jayanth and Liu, Gaowen and Wang, Zihan and Shang, Jingbo},
  booktitle={Proceedings of the 62nd Annual Meeting of the Association for Computational Linguistics (ACL)},
  pages={459--477},
  year={2024}
}

@inproceedings{coelho2024dwell,
  title={Dwell in the Beginning: How Language Models Embed Long Documents for Dense Retrieval},
  author={Coelho, Jo{\~a}o and Martins, Bruno and Magalh{\~a}es, Jo{\~a}o and Callan, Jamie and Xiong, Chenyan},
  booktitle={Proceedings of the 62nd Annual Meeting of the Association for Computational Linguistics (Volume 2: Short Papers)},
  pages={370--377},
  year={2024}
}

@article{man2024ullme,
  title={Ullme: A unified framework for large language model embeddings with generation-augmented learning},
  author={Man, Hieu and Ngo, Nghia Trung and Dernoncourt, Franck and Nguyen, Thien Huu},
  journal={arXiv preprint arXiv:2408.03402},
  year={2024}
}

@article{tan2025qaea,
author={Tan, Hongming and Zhan, Shaoxiong and Lin, Hai and Zheng, Hai-Tao and Chan, Wai Kin},
journal={IEEE Transactions on Knowledge \& Data Engineering (TKDE)},
title={{QAEA-DR: A Unified Text Augmentation Framework for Dense Retrieval}},
year={2025},
volume={37},
number={06},
ISSN={1558-2191},
pages={3669-3683},
doi={10.1109/TKDE.2025.3543203}
}

@article{li2024your,
  title={Your mixture-of-experts llm is secretly an embedding model for free},
  author={Li, Ziyue and Zhou, Tianyi},
  journal={arXiv preprint arXiv:2410.10814},
  year={2024}
}

@inproceedings{wang2024denosent,
  title={Denosent: A denoising objective for self-supervised sentence representation learning},
  author={Wang, Xinghao and He, Junliang and Wang, Pengyu and Zhou, Yunhua and Sun, Tianxiang and Qiu, Xipeng},
  booktitle={Proceedings of the AAAI Conference on Artificial Intelligence (AAAI)},
  volume={38},
  pages={19180--19188},
  year={2024}
}

@inproceedings{yoo2024hyper,
  title={Hyper-CL: Conditioning Sentence Representations with Hypernetworks},
  author={Yoo, Young and Cha, Jii and Kim, Changhyeon and Kim, Taeuk},
  booktitle={Proceedings of the 62nd Annual Meeting of the Association for Computational Linguistics (ACL)},
  pages={700--711},
  year={2024}
}

@article{tamber2024can,
  title={Can't Hide Behind the API: Stealing Black-Box Commercial Embedding Models},
  author={Tamber, Manveer Singh and Xian, Jasper and Lin, Jimmy},
  journal={arXiv preprint arXiv:2406.09355},
  year={2024}
}

@inproceedings{jeong2024simple,
  title={A Simple Angle-based Approach for Contrastive Learning of Unsupervised Sentence Representation},
  author={Jeong, Yoo Hyun and Han, Myeongsoo and Chae, Dong-Kyu},
  booktitle={Findings of the Association for Computational Linguistics: EMNLP 2024},
  pages={5553--5572},
  year={2024}
}

@article{jha2024jina,
  title={Jina-colbert-v2: A general-purpose multilingual late interaction retriever},
  author={Jha, Rohan and Wang, Bo and G{\"u}nther, Michael and Mastrapas, Georgios and Sturua, Saba and Mohr, Isabelle and Koukounas, Andreas and Akram, Mohammad Kalim and Wang, Nan and Xiao, Han},
  journal={arXiv preprint arXiv:2408.16672},
  year={2024}
}

@inproceedings{khattab2020colbert,
  title={Colbert: Efficient and effective passage search via contextualized late interaction over bert},
  author={Khattab, Omar and Zaharia, Matei},
  booktitle={Proceedings of the 43rd International ACM SIGIR conference on research and development in Information Retrieval (SIGIR)},
  pages={39--48},
  year={2020}
}

@inproceedings{thirukovalluru2024sumcse,
  title={SumCSE: Summary as a transformation for Contrastive Learning},
  author={Thirukovalluru, Raghuveer and Wang, Xiaolan and Chen, Jun and Li, Shuyang and Lei, Jie and Jin, Rong and Dhingra, Bhuwan},
  booktitle={Findings of the Association for Computational Linguistics: NAACL 2024},
  pages={3577--3588},
  year={2024}
}

@article{deng2025following,
  title={Following the Autoregressive Nature of LLM Embeddings via Compression and Alignment},
  author={Deng, Jingcheng and Jiang, Zhongtao and Pang, Liang and Chen, Liwei and Xu, Kun and Wei, Zihao and Shen, Huawei and Cheng, Xueqi},
  journal={arXiv preprint arXiv:2502.11401},
  year={2025}
}

@article{bhatia2024swan,
  title={Swan and arabicmteb: Dialect-aware, arabic-centric, cross-lingual, and cross-cultural embedding models and benchmarks},
  author={Bhatia, Gagan and Nagoudi, El Moatez Billah and Mekki, Abdellah El and Alwajih, Fakhraddin and Abdul-Mageed, Muhammad},
  journal={arXiv preprint arXiv:2411.01192},
  year={2024}
}

@article{thirukovalluru2024geneol,
  title={Geneol: Harnessing the generative power of llms for training-free sentence embeddings},
  author={Thirukovalluru, Raghuveer and Dhingra, Bhuwan},
  journal={arXiv preprint arXiv:2410.14635},
  year={2024}
}

@article{zhuang2024starbucks,
  title={Starbucks: Improved Training for 2D Matryoshka Embeddings},
  author={Zhuang, Shengyao and Wang, Shuai and Koopman, Bevan and Zuccon, Guido},
  journal={arXiv preprint arXiv:2410.13230},
  year={2024}
}

@inproceedings{xianmingese,
  title={ESE: Espresso Sentence Embeddings},
  author={Xianming, LI and Li, Zongxi and Li, Jing and Xie, Haoran and Li, Qing},
  booktitle={The Thirteenth International Conference on Learning Representations (ICLR)},
  year={2025}
}

@article{kusupati2022matryoshka,
  title={Matryoshka representation learning},
  author={Kusupati, Aditya and Bhatt, Gantavya and Rege, Aniket and Wallingford, Matthew and Sinha, Aditya and Ramanujan, Vivek and Howard-Snyder, William and Chen, Kaifeng and Kakade, Sham and Jain, Prateek and others},
  journal={Advances in Neural Information Processing Systems (NeurIPS)},
  volume={35},
  pages={30233--30249},
  year={2022}
}

@article{zhang2023language,
  title={Language models are universal embedders},
  author={Zhang, Xin and Li, Zehan and Zhang, Yanzhao and Long, Dingkun and Xie, Pengjun and Zhang, Meishan and Zhang, Min},
  journal={arXiv preprint arXiv:2310.08232},
  year={2023}
}

@article{moreira2024nv,
  title={NV-Retriever: Improving text embedding models with effective hard-negative mining},
  author={Moreira, Gabriel de Souza P and Osmulski, Radek and Xu, Mengyao and Ak, Ronay and Schifferer, Benedikt and Oldridge, Even},
  journal={arXiv preprint arXiv:2407.15831},
  year={2024}
}

@inproceedings{pappadopulo2024non,
  title={Non-contrastive sentence representations via self-supervision},
  author={Pappadopulo, Duccio and Farina, Marco},
  booktitle={Findings of the Association for Computational Linguistics: NAACL 2024},
  pages={4274--4284},
  year={2024}
}

@article{poswiata2024pl,
  title={PL-MTEB: Polish Massive Text Embedding Benchmark},
  author={Po{\'s}wiata, Rafa{\l} and Dadas, S{\l}awomir and Pere{\l}kiewicz, Micha{\l}},
  journal={arXiv preprint arXiv:2405.10138},
  year={2024}
}

@inproceedings{huang2023bridging,
  title={Bridging Continuous and Discrete Spaces: Interpretable Sentence Representation Learning via Compositional Operations},
  author={Huang, James and Yao, Wenlin and Song, Kaiqiang and Zhang, Hongming and Chen, Muhao and Yu, Dong},
  booktitle={Proceedings of the 2023 Conference on Empirical Methods in Natural Language Processing (EMNLP)},
  pages={14584--14595},
  year={2023},
}

@article{o2024disentangling,
  title={Disentangling dense embeddings with sparse autoencoders},
  author={O'Neill, Charles and Ye, Christine and Iyer, Kartheik and Wu, John F},
  journal={arXiv preprint arXiv:2408.00657},
  year={2024}
}

@inproceedings{cqgmbqa,
    title={A General Framework for Producing Interpretable Semantic Text Embeddings},
    author={Yiqun Sun and Qiang Huang and Yixuan Tang and Anthony K. H. Tung and Jun Yu},
    booktitle={The Thirteenth International Conference on Learning Representations (ICLR)},
    year={2025},
}

@article{xiao2025mieb,
  title={MIEB: Massive Image Embedding Benchmark},
  author={Xiao, Chenghao and Chung, Isaac and Kerboua, Imene and Stirling, Jamie and Zhang, Xin and Kardos, M{\'a}rton and Solomatin, Roman and Moubayed, Noura Al and Enevoldsen, Kenneth and Muennighoff, Niklas},
  journal={arXiv preprint arXiv:2504.10471},
  year={2025}
}

@inproceedings{sato2024improving,
  title={Improving Sentence Embeddings with Automatic Generation of Training Data Using Few-shot Examples},
  author={Sato, Soma and Tsukagoshi, Hayato and Sasano, Ryohei and Takeda, Koichi},
  booktitle={Proceedings of the 62nd Annual Meeting of the Association for Computational Linguistics (Volume 4: Student Research Workshop)},
  pages={519--530},
  year={2024}
}

@article{han2025ateb,
  title={ATEB: Evaluating and Improving Advanced NLP Tasks for Text Embedding Models},
  author={Han, Simeng and Gomez, Frank Palma and Vu, Tu and Li, Zefei and Cer, Daniel and Zeng, Hansi and Tar, Chris and Cohan, Arman and Abrego, Gustavo Hernandez},
  journal={arXiv preprint arXiv:2502.16766},
  year={2025}
}

@inproceedings{zhuang2024not,
  title={Not All Negatives are Equally Negative: Soft Contrastive Learning for Unsupervised Sentence Representations},
  author={Zhuang, Haojie and Emma Zhang, Wei and Yang, Jian and Chen, Weitong and Sheng, Quan Z},
  booktitle={Proceedings of the 33rd ACM International Conference on Information and Knowledge Management (CIKM)},
  pages={3591--3601},
  year={2024}
}

@article{tamber2025teaching,
  title={Teaching Dense Retrieval Models to Specialize with Listwise Distillation and LLM Data Augmentation},
  author={Tamber, Manveer Singh and Kazi, Suleman and Sourabh, Vivek and Lin, Jimmy},
  journal={arXiv preprint arXiv:2502.19712},
  year={2025}
}

@article{wang2025multi,
  title={Multi-Sense Embeddings for Language Models and Knowledge Distillation},
  author={Wang, Qitong and Zaki, Mohammed J and Kollias, Georgios and Kalantzis, Vasileios},
  journal={arXiv preprint arXiv:2504.06036},
  year={2025}
}

@article{lai2024enhancing,
  title={Enhancing Unsupervised Sentence Embeddings via Knowledge-Driven Data Augmentation and Gaussian-Decayed Contrastive Learning},
  author={Lai, Peichao and Zhang, Zhengfeng and Zhang, Wentao and Fu, Fangcheng and Cui, Bin},
  journal={arXiv preprint arXiv:2409.12887},
  year={2024}
}

@article{yu2024arctic,
  title={Arctic-Embed 2.0: Multilingual Retrieval Without Compromise},
  author={Yu, Puxuan and Merrick, Luke and Nuti, Gaurav and Campos, Daniel},
  journal={arXiv preprint arXiv:2412.04506},
  year={2024}
}

@article{xiao2024pixel,
  title={Pixel Sentence Representation Learning},
  author={Xiao, Chenghao and Huang, Zhuoxu and Chen, Danlu and Hudson, G Thomas and Li, Yizhi and Duan, Haoran and Lin, Chenghua and Fu, Jie and Han, Jungong and Moubayed, Noura Al},
  journal={arXiv preprint arXiv:2402.08183},
  year={2024}
}

@article{he2025refining,
  title={Refining Sentence Embedding Model through Ranking Sentences Generation with Large Language Models},
  author={He, Liyang and Liu, Chenglong and Li, Rui and Huang, Zhenya and Ruan, Shulan and Zhou, Jun and Chen, Enhong},
  journal={arXiv preprint arXiv:2502.13656},
  year={2025}
}

@inproceedings{zhao2025prompt,
  title={Prompt Tuning Can Simply Adapt Large Language Models to Text Encoders},
  author={Zhao, Kaiyan and Wu, Qiyu and Miao, Zhongtao and Tsuruoka, Yoshimasa},
  booktitle={Proceedings of the 10th Workshop on Representation Learning for NLP (RepL4NLP-2025)},
  pages={38--50},
  year={2025}
}

@article{banar2024beir,
  title={BEIR-NL: Zero-shot Information Retrieval Benchmark for the Dutch Language},
  author={Banar, Nikolay and Lotfi, Ehsan and Daelemans, Walter},
  journal={arXiv preprint arXiv:2412.08329},
  year={2024}
}

@article{ji2025learning,
  title={Learning More Effective Representations for Dense Retrieval through Deliberate Thinking Before Search},
  author={Ji, Yifan and Xu, Zhipeng and Liu, Zhenghao and Yan, Yukun and Yu, Shi and Li, Yishan and Liu, Zhiyuan and Gu, Yu and Yu, Ge and Sun, Maosong},
  journal={arXiv preprint arXiv:2502.12974},
  year={2025}
}

@article{kasmaee2024chemteb,
  title={ChemTEB: Chemical Text Embedding Benchmark, an Overview of Embedding Models Performance \& Efficiency on a Specific Domain},
  author={Kasmaee, Ali Shiraee and Khodadad, Mohammad and Saloot, Mohammad Arshi and Sherck, Nicholas and Dokas, Stephen and Mahyar, Hamidreza and Samiee, Soheila},
  journal={arXiv preprint arXiv:2412.00532},
  year={2024}
}

@article{zinvandi2025famteb,
  title={FaMTEB: Massive Text Embedding Benchmark in Persian Language},
  author={Zinvandi, Erfan and Alikhani, Morteza and Sarmadi, Mehran and Pourbahman, Zahra and Arvin, Sepehr and Kazemi, Reza and Amini, Arash},
  journal={arXiv preprint arXiv:2502.11571},
  year={2025}
}

@article{ananthakrishnan2025can,
  title={Can Cross Encoders Produce Useful Sentence Embeddings?},
  author={Ananthakrishnan, Haritha and Dolby, Julian and Kokel, Harsha and Samulowitz, Horst and Srinivas, Kavitha},
  journal={arXiv preprint arXiv:2502.03552},
  year={2025}
}

@article{li2024improving,
  title={Improving General Text Embedding Model: Tackling Task Conflict and Data Imbalance through Model Merging},
  author={Li, Mingxin and Nie, Zhijie and Zhang, Yanzhao and Long, Dingkun and Zhang, Richong and Xie, Pengjun},
  journal={arXiv preprint arXiv:2410.15035},
  year={2024}
}

@article{tang2025finmteb,
  title={FinMTEB: Finance Massive Text Embedding Benchmark},
  author={Tang, Yixuan and Yang, Yi},
  journal={arXiv preprint arXiv:2502.10990},
  year={2025}
}

@article{fu2024token,
  title={Token Prepending: A Training-Free Approach for Eliciting Better Sentence Embeddings from LLMs},
  author={Fu, Yuchen and Cheng, Zifeng and Jiang, Zhiwei and Wang, Zhonghui and Yin, Yafeng and Li, Zhengliang and Gu, Qing},
  journal={arXiv preprint arXiv:2412.11556},
  year={2024}
}

@article{yamada2025out,
  title={Out-of-the-Box Conditional Text Embeddings from Large Language Models},
  author={Yamada, Kosuke and Zhang, Peinan},
  journal={arXiv preprint arXiv:2504.16411},
  year={2025}
}

@article{kovalev2025building,
  title={Building Russian Benchmark for Evaluation of Information Retrieval Models},
  author={Kovalev, Grigory and Tikhomirov, Mikhail and Kozhevnikov, Evgeny and Kornilov, Max and Loukachevitch, Natalia},
  journal={arXiv preprint arXiv:2504.12879},
  year={2025}
}

@inproceedings{ponwitayarat2024space,
  title={Space Decomposition for Sentence Embedding},
  author={Ponwitayarat, Wuttikorn and Limkonchotiwat, Peerat and Chuangsuwanich, Ekapol and Nutanong, Sarana},
  booktitle={Findings of the Association for Computational Linguistics: ACL 2024},
  pages={11227--11239},
  year={2024}
}

@article{zhang2025cse,
  title={CSE-SFP: Enabling Unsupervised Sentence Representation Learning via a Single Forward Pass},
  author={Zhang, Bowen and Song, Zixin and Li, Chunping},
  journal={arXiv preprint arXiv:2505.00389},
  year={2025}
}

@inproceedings{sturua2025jina,
  title={Jina Embeddings V3: Multilingual Text Encoder with Low-Rank Adaptations},
  author={Sturua, Saba and Mohr, Isabelle and Kalim Akram, Mohammad and G{\"u}nther, Michael and Wang, Bo and Krimmel, Markus and Wang, Feng and Mastrapas, Georgios and Koukounas, Andreas and Wang, Nan and others},
  booktitle={European Conference on Information Retrieval (ECIR)},
  pages={123--129},
  year={2025}
}

@article{gill2025advancing,
  title={Advancing Semantic Caching for LLMs with Domain-Specific Embeddings and Synthetic Data},
  author={Gill, Waris and Cechmanek, Justin and Hutcherson, Tyler and Rajamohan, Srijith and Agarwal, Jen and Gulzar, Muhammad Ali and Singh, Manvinder and Dion, Benoit},
  journal={arXiv preprint arXiv:2504.02268},
  year={2025}
}

@inproceedings{thakur2beir,
  title={{BEIR:} A Heterogeneous Benchmark for Zero-shot Evaluation of Information Retrieval Models},
  author={Thakur, Nandan and Reimers, Nils and R{\"u}ckl{\'e}, Andreas and Srivastava, Abhishek and Gurevych, Iryna},
  booktitle={Proceedings of the Neural Information Processing Systems Track on Datasets and Benchmarks},
  year = "2021"
}

@inproceedings{jha2018interpretable,
  title={Interpretable word embeddings for medical domain},
  author={Jha, Kishlay and Wang, Yaqing and Xun, Guangxu and Zhang, Aidong},
  booktitle={2018 IEEE International Conference on Data Mining (ICDM)},
  pages={1061--1066},
  year={2018},
}

@article{senel2018semantic,
  title={Semantic Structure and Interpretability of Word Embeddings},
  author={Senel, Lutfi Kerem and Utlu, Ihsan and Yucesoy, Veysel and Koc, Aykut and Cukur, Tolga},
  journal={IEEE/ACM Transactions on Audio, Speech and Language Processing (TASLP)},
  volume={26},
  number={10},
  pages={1769--1779},
  year={2018},
}

@inproceedings{subramanian2018spine,
  title={{SPINE:} {SP}arse {I}nterpretable {N}eural {E}mbeddings},
  author={Subramanian, Anant and Pruthi, Danish and Jhamtani, Harsh and Berg-Kirkpatrick, Taylor and Hovy, Eduard},
  booktitle={Proceedings of the AAAI Conference on Artificial Intelligence (AAAI)},
  pages={4921--4928},
  year={2018}
}

@inproceedings{panigrahi-etal-2019-word2sense,
    title = "{W}ord2{S}ense: Sparse Interpretable Word Embeddings",
    author = "Panigrahi, Abhishek  and
      Simhadri, Harsha Vardhan  and
      Bhattacharyya, Chiranjib",
    booktitle = "Proceedings of the 57th Annual Meeting of the Association for Computational Linguistics (ACL)",
    year = "2019",
    pages = "5692--5705",
}

@inproceedings{opitz-frank-2022-sbert,
    title = "{SBERT} studies Meaning Representations: Decomposing Sentence Embeddings into Explainable Semantic Features",
    author = "Opitz, Juri  and
      Frank, Anette",
    booktitle = "Proceedings of the 2nd Conference of the Asia-Pacific Chapter of the Association for Computational Linguistics and the 12th International Joint Conference on Natural Language Processing (AACL-IJCNLP)",
    year = "2022",
    pages = "625--638",
}

@inproceedings{simhi-markovitch-2023-interpreting,
    title = "Interpreting Embedding Spaces by Conceptualization",
    author = "Simhi, Adi  and
      Markovitch, Shaul",
    booktitle = "Proceedings of the 2023 Conference on Empirical Methods in Natural Language Processing (EMNLP)",
    year = "2023",
    pages = "1704--1719",
}

@inproceedings{mcinerney-etal-2023-chill,
    title = "{CH}i{LL}: Zero-shot Custom Interpretable Feature Extraction from Clinical Notes with Large Language Models",
    author = "McInerney, Denis  and
      Young, Geoffrey  and
      van de Meent, Jan-Willem  and
      Wallace, Byron",
    booktitle = "Findings of the Association for Computational Linguistics: EMNLP 2023",
    year = "2023",
    pages = "8477--8494",
}

@inproceedings{
Benara2024CraftingIE,
title={Crafting Interpretable Embeddings for Language Neuroscience by Asking {LLM}s Questions},
author={Vinamra Benara and Chandan Singh and John Xavier Morris and Richard Antonello and Ion Stoica and Alexander Huth and Jianfeng Gao},
booktitle={The Thirty-eighth Annual Conference on Neural Information Processing Systems (NeurIPS)},
volume={37},
pages={124137},
year={2024},
}

@inproceedings{santhanam2022colbertv2,
  title={ColBERTv2: Effective and Efficient Retrieval via Lightweight Late Interaction},
  author={Santhanam, Keshav and Khattab, Omar and Saad-Falcon, Jon and Potts, Christopher and Zaharia, Matei},
  booktitle={Proceedings of the 2022 Conference of the North American Chapter of the Association for Computational Linguistics: Human Language Technologies (NAACL-HLT)},
  pages={3715--3734},
  year={2022}
}

@article{liu2019roberta,
  title={{RoBERTa}: A Robustly Optimized {BERT} Pretraining Approach},
  author={Liu, Yinhan and Ott, Myle and Goyal, Naman and Du, Jingfei and Joshi, Mandar and Chen, Danqi and Levy, Omer and Lewis, Mike and Zettlemoyer, Luke and Stoyanov, Veselin},
  journal={arXiv preprint arXiv:1907.11692},
  year={2019}
}

@inproceedings{sts14,
    title = "{S}em{E}val-2014 Task 10: Multilingual Semantic Textual Similarity",
    author = "Agirre, Eneko  and
      Banea, Carmen  and
      Cardie, Claire  and
      Cer, Daniel  and
      Diab, Mona  and
      Gonzalez-Agirre, Aitor  and
      Guo, Weiwei  and
      Mihalcea, Rada  and
      Rigau, German  and
      Wiebe, Janyce",
    booktitle = "Proceedings of the 8th International Workshop on Semantic Evaluation ({S}em{E}val 2014)",
    year = "2014",
    pages = "81--91",
}

@inproceedings{sts15,
    title = "{S}em{E}val-2015 Task 2: Semantic Textual Similarity, {E}nglish, {S}panish and Pilot on Interpretability",
    author = "Agirre, Eneko  and
      Banea, Carmen  and
      Cardie, Claire  and
      Cer, Daniel  and
      Diab, Mona  and
      Gonzalez-Agirre, Aitor  and
      Guo, Weiwei  and
      Lopez-Gazpio, I{\~n}igo  and
      Maritxalar, Montse  and
      Mihalcea, Rada  and
      Rigau, German  and
      Uria, Larraitz  and
      Wiebe, Janyce",
    booktitle = "Proceedings of the 9th International Workshop on Semantic Evaluation ({S}em{E}val 2015)",
    year = "2015",
    pages = "252--263",
}

@inproceedings{sts16,
    title = "{S}em{E}val-2016 Task 1: Semantic Textual Similarity, Monolingual and Cross-Lingual Evaluation",
    author = "Agirre, Eneko  and
      Banea, Carmen  and
      Cer, Daniel  and
      Diab, Mona  and
      Gonzalez-Agirre, Aitor  and
      Mihalcea, Rada  and
      Rigau, German  and
      Wiebe, Janyce",
    booktitle = "Proceedings of the 10th International Workshop on Semantic Evaluation ({S}em{E}val-2016)",
    year = "2016",
    pages = "497--511",
}

@inproceedings{sts12,
    title = "{S}em{E}val-2012 Task 6: A Pilot on Semantic Textual Similarity",
    author = "Agirre, Eneko  and
      Cer, Daniel  and
      Diab, Mona  and
      Gonzalez-Agirre, Aitor",
    booktitle = "*{SEM} 2012: The First Joint Conference on Lexical and Computational Semantics {--} Volume 1: Proceedings of the main conference and the shared task, and Volume 2: Proceedings of the Sixth International Workshop on Semantic Evaluation ({S}em{E}val 2012)",
    year = "2012",
    url = "https://aclanthology.org/S12-1051/",
    pages = "385--393"
}

@inproceedings{sts13,
    title = "*{SEM} 2013 shared task: Semantic Textual Similarity",
    author = "Agirre, Eneko  and
      Cer, Daniel  and
      Diab, Mona  and
      Gonzalez-Agirre, Aitor  and
      Guo, Weiwei",
    booktitle = "Second Joint Conference on Lexical and Computational Semantics (*{SEM}), Volume 1: Proceedings of the Main Conference and the Shared Task: Semantic Textual Similarity",
    month = jun,
    year = "2013",
}

@article{sts17,
  title={Semeval-2017 task 1: Semantic textual similarity-multilingual and cross-lingual focused evaluation},
  author={Cer, Daniel and Diab, Mona and Agirre, Eneko and Lopez-Gazpio, Inigo and Specia, Lucia},
  journal={arXiv preprint arXiv:1708.00055},
  year={2017}
}

@inproceedings{dolan2005automatically,
  title={Automatically constructing a corpus of sentential paraphrases},
  author={Dolan, Bill and Brockett, Chris},
  booktitle={Third international workshop on paraphrasing (IWP2005)},
  year={2005}
}

@article{sharma2019natural,
  title={Natural language understanding with the quora question pairs dataset},
  author={Sharma, Lakshay and Graesser, Laura and Nangia, Nikita and Evci, Utku},
  journal={arXiv preprint arXiv:1907.01041},
  year={2019}
}

@inproceedings{sick-r,
    title = "A {SICK} cure for the evaluation of compositional distributional semantic models",
    author = "Marelli, Marco  and
      Menini, Stefano  and
      Baroni, Marco  and
      Bentivogli, Luisa  and
      Bernardi, Raffaella  and
      Zamparelli, Roberto",
    booktitle = "Proceedings of the Ninth International Conference on Language Resources and Evaluation (LREC)",
    year = "2014",
    pages = "216--223",
}

@misc{huggingface:dataset:stsb_multi_mt,
  author = {Philip May},
  title = {Machine translated multilingual STS benchmark dataset.},
  url = {https://github.com/PhilipMay/stsb-multi-mt},
  year = {2021},
}

@inproceedings{snli,
    title = "A large annotated corpus for learning natural language inference",
    author = "Bowman, Samuel R.  and
      Angeli, Gabor  and
      Potts, Christopher  and
      Manning, Christopher D.",
    booktitle = "Proceedings of the 2015 Conference on Empirical Methods in Natural Language Processing (EMNLP)",
    year = "2015",
    pages = "632--642",
}

@inproceedings{williams2018broad,
  title={A broad-coverage challenge corpus for sentence understanding through inference},
  author={Williams, Adina and Nangia, Nikita and Bowman, Samuel R},
  booktitle={2018 Conference of the North American Chapter of the Association for Computational Linguistics: Human Language Technologies, NAACL HLT 2018},
  pages={1112--1122},
  year={2018},
}

@inproceedings{nie2020adversarial,
  title={Adversarial NLI: A New Benchmark for Natural Language Understanding},
  author={Nie, Yixin and Williams, Adina and Dinan, Emily and Bansal, Mohit and Weston, Jason and Kiela, Douwe},
  booktitle={Proceedings of the 58th Annual Meeting of the Association for Computational Linguistics (ACL)},
  pages={4885--4901},
  year={2020}
}

@article{demszky2018transforming,
  title={Transforming question answering datasets into natural language inference datasets},
  author={Demszky, Dorottya and Guu, Kelvin and Liang, Percy},
  journal={arXiv preprint arXiv:1809.02922},
  year={2018}
}

@article{rajpurkar2016squad,
  title={Squad: 100,000+ questions for machine comprehension of text},
  author={Rajpurkar, Pranav and Zhang, Jian and Lopyrev, Konstantin and Liang, Percy},
  journal={arXiv preprint arXiv:1606.05250},
  year={2016}
}

@article{zhang2023miracl,
  title={Miracl: A multilingual retrieval dataset covering 18 diverse languages},
  author={Zhang, Xinyu and Thakur, Nandan and Ogundepo, Odunayo and Kamalloo, Ehsan and Alfonso-Hermelo, David and Li, Xiaoguang and Liu, Qun and Rezagholizadeh, Mehdi and Lin, Jimmy},
  journal={Transactions of the Association for Computational Linguistics (TACL)},
  volume={11},
  pages={1114--1131},
  year={2023},
}

@inproceedings{zhang-etal-2021-mr,
    title = "Mr. {T}y{D}i: A Multi-lingual Benchmark for Dense Retrieval",
    author = "Zhang, Xinyu  and
      Ma, Xueguang  and
      Shi, Peng  and
      Lin, Jimmy",
    booktitle = "Proceedings of the 1st Workshop on Multilingual Representation Learning",
    year = "2021",
    doi = "10.18653/v1/2021.mrl-1.12",
    pages = "127--137",
}

@inproceedings{wang2013theoretical,
  title={A Theoretical Analysis of {NDCG} Type Ranking Measures},
  author={Wang, Yining and Wang, Liwei and Li, Yuanzhi and He, Di and Liu, Tie-Yan},
  booktitle={Proceedings of the 26th Annual Conference on Learning Theory (COLT)},
  pages={25--54},
  year={2013}
}

@inproceedings{xiong2024benchmarking,
  title={Benchmarking retrieval-augmented generation for medicine},
  author={Xiong, Guangzhi and Jin, Qiao and Lu, Zhiyong and Zhang, Aidong},
  booktitle={Findings of the Association for Computational Linguistics: ACL 2024},
  pages={6233--6251},
  year={2024}
}

@inproceedings{chen2024benchmarking,
  title={Benchmarking large language models in retrieval-augmented generation},
  author={Chen, Jiawei and Lin, Hongyu and Han, Xianpei and Sun, Le},
  booktitle={Proceedings of the AAAI Conference on Artificial Intelligence (AAAI)},
  pages={17754--17762},
  year={2024}
}

@article{friel2024ragbench,
  title={Ragbench: Explainable benchmark for retrieval-augmented generation systems},
  author={Friel, Robert and Belyi, Masha and Sanyal, Atindriyo},
  journal={arXiv preprint arXiv:2407.11005},
  year={2024}
}

@inproceedings{niu2024ragtruth,
  title={RAGTruth: A Hallucination Corpus for Developing Trustworthy Retrieval-Augmented Language Models},
  author={Niu, Cheng and Wu, Yuanhao and Zhu, Juno and Xu, Siliang and Shum, Kashun and Zhong, Randy and Song, Juntong and Zhang, Tong},
  booktitle={Proceedings of the 62nd Annual Meeting of the Association for Computational Linguistics (ACL)},
  pages={10862--10878},
  year={2024}
}

@article{wang2024domainrag,
  title={Domainrag: A chinese benchmark for evaluating domain-specific retrieval-augmented generation},
  author={Wang, Shuting and Liu, Jiongnan and Song, Shiren and Cheng, Jiehan and Fu, Yuqi and Guo, Peidong and Fang, Kun and Zhu, Yutao and Dou, Zhicheng},
  journal={arXiv preprint arXiv:2406.05654},
  year={2024}
}

@inproceedings{rau2024bergen,
  title={BERGEN: A Benchmarking Library for Retrieval-Augmented Generation},
  author={Rau, David and D{\'e}jean, Herv{\'e} and Chirkova, Nadezhda and Formal, Thibault and Wang, Shuai and Clinchant, St{\'e}phane and Nikoulina, Vassilina},
  booktitle={Findings of the Association for Computational Linguistics: EMNLP 2024},
  pages={7640--7663},
  year={2024}
}

@article{tang2024multihop,
  title={Multihop-rag: Benchmarking retrieval-augmented generation for multi-hop queries},
  author={Tang, Yixuan and Yang, Yi},
  journal={arXiv preprint arXiv:2401.15391},
  year={2024}
}

@article{pipitone2024legalbench,
  title={Legalbench-rag: A benchmark for retrieval-augmented generation in the legal domain},
  author={Pipitone, Nicholas and Alami, Ghita Houir},
  journal={arXiv preprint arXiv:2408.10343},
  year={2024}
}

@inproceedings{tihanyi2024cybermetric,
  title={CyberMetric: a benchmark dataset based on retrieval-augmented generation for evaluating LLMs in cybersecurity knowledge},
  author={Tihanyi, Norbert and Ferrag, Mohamed Amine and Jain, Ridhi and Bisztray, Tamas and Debbah, Merouane},
  booktitle={2024 IEEE International Conference on Cyber Security and Resilience (CSR)},
  pages={296--302},
  year={2024},
}

@article{hui2024uda,
  title={Uda: A benchmark suite for retrieval augmented generation in real-world document analysis},
  author={Hui, Yulong and Lu, Yao and Zhang, Huanchen},
  journal={arXiv preprint arXiv:2406.15187},
  year={2024}
}

@inproceedings{salemi2024evaluating,
  title={Evaluating retrieval quality in retrieval-augmented generation},
  author={Salemi, Alireza and Zamani, Hamed},
  booktitle={Proceedings of the 47th International ACM SIGIR Conference on Research and Development in Information Retrieval (SIGIR)},
  pages={2395--2400},
  year={2024}
}

@inproceedings{lewis2020retrieval,
  title={Retrieval-{A}ugmented {G}eneration for {K}nowledge-{I}ntensive {NLP} {T}asks},
  author={Lewis, Patrick and Perez, Ethan and Piktus, Aleksandra and Petroni, Fabio and Karpukhin, Vladimir and Goyal, Naman and K{\"u}ttler, Heinrich and Lewis, Mike and Yih, Wen-tau and Rockt{\"a}schel, Tim and others},
  booktitle={Proceedings of the 34th International Conference on Neural Information Processing Systems (NeurIPS)},
  pages={9459--9474},
  year={2020}
}

@inproceedings{park2025mirage,
  title={MIRAGE: A Metric-Intensive Benchmark for Retrieval-Augmented Generation Evaluation},
  author={Park, Chanhee and Moon, Hyeonseok and Park, Chanjun and Lim, Heui-Seok},
  booktitle={Findings of the Association for Computational Linguistics: NAACL 2025},
  pages={2883--2900},
  year={2025}
}

@article{chen2024bge,
  title={Bge m3-embedding: Multi-lingual, multi-functionality, multi-granularity text embeddings through self-knowledge distillation},
  author={Chen, Jianlv and Xiao, Shitao and Zhang, Peitian and Luo, Kun and Lian, Defu and Liu, Zheng},
  journal={arXiv preprint arXiv:2402.03216},
  year={2024}
}

@article{mohr2024multi,
  title={Multi-task contrastive learning for 8192-token bilingual text embeddings},
  author={Mohr, Isabelle and Krimmel, Markus and Sturua, Saba and Akram, Mohammad Kalim and Koukounas, Andreas and G{\"u}nther, Michael and Mastrapas, Georgios and Ravishankar, Vinit and Mart{\'\i}nez, Joan Fontanals and Wang, Feng and others},
  journal={arXiv preprint arXiv:2402.17016},
  year={2024}
}

@inproceedings{gunther-etal-2023-jina,
    title = "{J}ina Embeddings: A Novel Set of High-Performance Sentence Embedding Models",
    author = {G{\"u}nther, Michael  and
      Milliken, Louis  and
      Geuter, Jonathan  and
      Mastrapas, Georgios  and
      Wang, Bo  and
      Xiao, Han},
    booktitle = "Proceedings of the 3rd Workshop for Natural Language Processing Open Source Software (NLP-OSS 2023)",
    month = dec,
    year = "2023",
}

@article{gunther2023jina,
  title={Jina embeddings 2: 8192-token general-purpose text embeddings for long documents},
  author={G{\"u}nther, Michael and Ong, Jackmin and Mohr, Isabelle and Abdessalem, Alaeddine and Abel, Tanguy and Akram, Mohammad Kalim and Guzman, Susana and Mastrapas, Georgios and Sturua, Saba and Wang, Bo and others},
  journal={arXiv preprint arXiv:2310.19923},
  year={2023}
}

@inproceedings{muennighoff2024generative,
  title={Generative representational instruction tuning},
  author={Muennighoff, Niklas and Hongjin, SU and Wang, Liang and Yang, Nan and Wei, Furu and Yu, Tao and Singh, Amanpreet and Kiela, Douwe},
  booktitle={ICLR 2024 Workshop: How Far Are We From AGI},
  year={2024}
}

@article{weller2024followir,
  title={Followir: Evaluating and teaching information retrieval models to follow instructions},
  author={Weller, Orion and Chang, Benjamin and MacAvaney, Sean and Lo, Kyle and Cohan, Arman and Van Durme, Benjamin and Lawrie, Dawn and Soldaini, Luca},
  journal={arXiv preprint arXiv:2403.15246},
  year={2024}
}

@article{yang2018hotpotqa,
  title={HotpotQA: A dataset for diverse, explainable multi-hop question answering},
  author={Yang, Zhilin and Qi, Peng and Zhang, Saizheng and Bengio, Yoshua and Cohen, William W and Salakhutdinov, Ruslan and Manning, Christopher D},
  journal={arXiv preprint arXiv:1809.09600},
  year={2018}
}

@inproceedings{peters-etal-2018-deep,
    title = "Deep Contextualized Word Representations",
    author = "Peters, Matthew E.  and
      Neumann, Mark  and
      Iyyer, Mohit  and
      Gardner, Matt  and
      Clark, Christopher  and
      Lee, Kenton  and
      Zettlemoyer, Luke",
    booktitle = "Proceedings of the 2018 Conference of the North {A}merican Chapter of the Association for Computational Linguistics: Human Language Technologies (NAACL-HLT)",
    year = "2018",
    pages = "2227--2237",
}

@article{sharifymoghaddam2025chatbot,
  title={Chatbot Arena Meets Nuggets: Towards Explanations and Diagnostics in the Evaluation of LLM Responses},
  author={Sharifymoghaddam, Sahel and Upadhyay, Shivani and Thakur, Nandan and Pradeep, Ronak and Lin, Jimmy},
  journal={arXiv preprint arXiv:2504.20006},
  year={2025}
}

@article{raffel2020exploring,
  title={Exploring the limits of transfer learning with a unified text-to-text transformer},
  author={Raffel, Colin and Shazeer, Noam and Roberts, Adam and Lee, Katherine and Narang, Sharan and Matena, Michael and Zhou, Yanqi and Li, Wei and Liu, Peter J},
  journal={Journal of Machine Learning Research (JMLR)},
  volume={21},
  number={140},
  pages={1--67},
  year={2020}
}

@article{chen2024little,
  title={Little Giants: Synthesizing High-Quality Embedding Data at Scale},
  author={Chen, Haonan and Wang, Liang and Yang, Nan and Zhu, Yutao and Zhao, Ziliang and Wei, Furu and Dou, Zhicheng},
  journal={arXiv preprint arXiv:2410.18634},
  year={2024}
}

@inproceedings{sravanthi2024pub,
  title={PUB: A Pragmatics Understanding Benchmark for Assessing LLMs’ Pragmatics Capabilities},
  author={Sravanthi, Settaluri and Doshi, Meet and Tankala, Pavan and Murthy, Rudra and Dabre, Raj and Bhattacharyya, Pushpak},
  booktitle={Findings of the Association for Computational Linguistics: ACL 2024},
  pages={12075--12097},
  year={2024}
}

@inproceedings{yue2024large,
  title={Do large language models understand conversational implicature--a case study with a Chinese sitcom},
  author={Yue, Shisen and Song, Siyuan and Cheng, Xinyuan and Hu, Hai},
  booktitle={China National Conference on Chinese Computational Linguistics},
  pages={402--418},
  year={2024},
}

@article{wei2025semantic,
  title={Semantic-KG: Using Knowledge Graphs to Construct Benchmarks for Measuring Semantic Similarity},
  author={Wei, Qiyao and Morrell, Edward and Goetz, Lea and van der Schaar, Mihaela},
  journal={arXiv preprint arXiv:2511.19925},
  year={2025}
}

@article{kazemi2023boardgameqa,
  title={Boardgameqa: A dataset for natural language reasoning with contradictory information},
  author={Kazemi, Mehran and Yuan, Quan and Bhatia, Deepti and Kim, Najoung and Xu, Xin and Imbrasaite, Vaiva and Ramachandran, Deepak},
  journal={Advances in Neural Information Processing Systems (NeurIPS)},
  volume={36},
  pages={39052--39074},
  year={2023}
}

@article{li2023diplomat,
  title={DiPlomat: a dialogue dataset for situated pragmatic reasoning},
  author={Li, Hengli and Zhu, Song-Chun and Zheng, Zilong},
  journal={Advances in Neural Information Processing Systems (NeurIPS)},
  volume={36},
  pages={46856--46884},
  year={2023}
}

@inproceedings{li2020molweni,
  title={Molweni: A Challenge Multiparty Dialogues-based Machine Reading Comprehension Dataset with Discourse Structure},
  author={Li, Jiaqi and Liu, Ming and Kan, Min-Yen and Zheng, Zihao and Wang, Zekun and Lei, Wenqiang and Liu, Ting and Qin, Bing},
  booktitle={Proceedings of the 28th International Conference on Computational Linguistics (COLING)},
  pages={2642--2652},
  year={2020}
}

@inproceedings{curry2024classist,
  title={Classist Tools: Social Class Correlates with Performance in NLP},
  author={Curry, Amanda Cercas and Attanasio, Giuseppe and Talat, Zeerak and Hovy, Dirk},
  booktitle={Proceedings of the 62nd Annual Meeting of the Association for Computational Linguistics (ACL)},
  pages={12643--12655},
  year={2024}
}

@inproceedings{baly2020we,
  title={We Can Detect Your Bias: Predicting the Political Ideology of News Articles},
  author={Baly, Ramy and Da San Martino, Giovanni and Glass, James and Nakov, Preslav},
  booktitle={Proceedings of the 2020 Conference on Empirical Methods in Natural Language Processing (EMNLP)},
  pages={4982--4991},
  year={2020}
}

@inproceedings{sap2020social,
  title={Social Bias Frames: Reasoning about Social and Power Implications of Language},
  author={Sap, Maarten and Gabriel, Saadia and Qin, Lianhui and Jurafsky, Dan and Smith, Noah A and Choi, Yejin},
  booktitle={Proceedings of the 58th Annual Meeting of the Association for Computational Linguistics (ACL)},
  pages={5477--5490},
  year={2020}
}

@inproceedings{elsherief2021latent,
  title={Latent Hatred: A Benchmark for Understanding Implicit Hate Speech},
  author={ElSherief, Mai and Ziems, Caleb and Muchlinski, David and Anupindi, Vaishnavi and Seybolt, Jordyn and De Choudhury, Munmun and Yang, Diyi},
  booktitle={Proceedings of the 2021 Conference on Empirical Methods in Natural Language Processing (EMNLP)},
  pages={345--363},
  year={2021}
}

@inproceedings{li2021p,
  title={P-stance: A large dataset for stance detection in political domain},
  author={Li, Yingjie and Sosea, Tiberiu and Sawant, Aditya and Nair, Ajith Jayaraman and Inkpen, Diana and Caragea, Cornelia},
  booktitle={Findings of the Association for Computational Linguistics: ACL 2021},
  pages={2355--2365},
  year={2021}
}

@inproceedings{louis2020d,
  title={“I’d rather just go to bed”: Understanding Indirect Answers},
  author={Louis, Annie and Roth, Dan and Radlinski, Filip},
  booktitle={Proceedings of the 2020 Conference on Empirical Methods in Natural Language Processing (EMNLP)},
  pages={7411--7425},
  year={2020}
}

@inproceedings{zheng2021grice,
  title={Grice: A grammar-based dataset for recovering implicature and conversational reasoning},
  author={Zheng, Zilong and Qiu, Shuwen and Fan, Lifeng and Zhu, Yixin and Zhu, Song-Chun},
  booktitle={Findings of the Association for Computational Linguistics: ACL 2021},
  pages={2074--2085},
  year={2021}
}

@inproceedings{liu2022testing,
  title={Testing the Ability of Language Models to Interpret Figurative Language},
  author={Liu, Emmy and Cui, Chenxuan and Zheng, Kenneth and Neubig, Graham},
  booktitle={Proceedings of the 2022 Conference of the North American Chapter of the Association for Computational Linguistics: Human Language Technologies (NAACL-HLT)},
  year={2022},
  pages = "4437--4452",
}

@inproceedings{chakrabarty2022flute,
  title={FLUTE: Figurative Language Understanding through Textual Explanations},
  author={Chakrabarty, Tuhin and Saakyan, Arkadiy and Ghosh, Debanjan and Muresan, Smaranda},
  booktitle={Proceedings of the 2022 Conference on Empirical Methods in Natural Language Processing (EMNLP)},
  pages={7139--7159},
  year={2022}
}

@inproceedings{jeretic2020natural,
  title={Are Natural Language Inference Models IMPPRESsive? Learning IMPlicature and PRESupposition},
  author={Jeretic, Paloma and Warstadt, Alex and Bhooshan, Suvrat and Williams, Adina},
  booktitle={Proceedings of the 58th Annual Meeting of the Association for Computational Linguistics (ACL)},
  pages={4437--4452},
  year={2020}
}

@inproceedings{parrish2021nope,
  title={NOPE: A Corpus of Naturally-Occurring Presuppositions in English},
  author={Parrish, Alicia and Schuster, Sebastian and Warstadt, Alex and Agha, Omar and Lee, Soo-Hwan and Zhao, Zhuoye and Bowman, Samuel and Linzen, Tal},
  booktitle={Proceedings of the 25th Conference on Computational Natural Language Learning (CoNLL)},
  pages={349--366},
  year={2021}
}

@article{solatorio2024gistembed,
  title={Gistembed: Guided in-sample selection of training negatives for text embedding fine-tuning},
  author={Solatorio, Aivin V},
  journal={arXiv preprint arXiv:2402.16829},
  year={2024}
}

@misc{emb2024mxbai,
  title={Open Source Strikes Bread - New Fluffy Embeddings Model},
  author={Sean Lee and Aamir Shakir and Darius Koenig and Julius Lipp},
  year={2024},
  url={https://www.mixedbread.ai/blog/mxbai-embed-large-v1},
}

@inproceedings{bge_embedding,
  title={C-pack: Packed resources for general chinese embeddings},
  author={Xiao, Shitao and Liu, Zheng and Zhang, Peitian and Muennighoff, Niklas and Lian, Defu and Nie, Jian-Yun},
  booktitle={Proceedings of the 47th international ACM SIGIR conference on research and development in information retrieval (SIGIR)},
  pages={641--649},
  year={2024}
}

@article{zhang2025jasperstelladistillationsota,
  title={Jasper and Stella: distillation of SOTA embedding models},
  author={Zhang, Dun and Li, Jiacheng and Zeng, Ziyang and Wang, Fulong},
  journal={arXiv preprint arXiv:2412.19048},
  year={2024}
}

@misc{LinqAIResearch2024,
  title={Linq-Embed-Mistral:Elevating Text Retrieval with Improved GPT Data Through Task-Specific Control and Quality Refinement},
  author = {Junseong Kim and Seolhwa Lee and Jihoon Kwon and Sangmo Gu and Yejin Kim and Minkyung Cho and Jy-yong Sohn and Chanyeol Choi},
  howpublished={Linq AI Research Blog},
  year={2024},
  url={https://getlinq.com/blog/linq-embed-mistral/}
}

@article{li2023towards,
  title={Towards general text embeddings with multi-stage contrastive learning},
  author={Li, Zehan and Zhang, Xin and Zhang, Yanzhao and Long, Dingkun and Xie, Pengjun and Zhang, Meishan},
  journal={arXiv preprint arXiv:2308.03281},
  year={2023}
}

@inproceedings{angelov2024topic,
  title={Topic Modeling: Contextual Token Embeddings Are All You Need},
  author={Angelov, Dimo and Inkpen, Diana},
  booktitle={Findings of the Association for Computational Linguistics: EMNLP 2024},
  pages={13528--13539},
  year={2024}
}

@article{grootendorst2022bertopic,
  title={BERTopic: Neural topic modeling with a class-based TF-IDF procedure},
  author={Grootendorst, Maarten},
  journal={arXiv preprint arXiv:2203.05794},
  year={2022}
}

@inproceedings{karpukhin2020dense,
  title={Dense Passage Retrieval for Open-Domain Question Answering},
  author={Karpukhin, Vladimir and Oguz, Barlas and Min, Sewon and Lewis, Patrick and Wu, Ledell and Edunov, Sergey and Chen, Danqi and Yih, Wen-tau},
  booktitle={Proceedings of the 2020 Conference on Empirical Methods in Natural Language Processing (EMNLP)},
  pages={6769--6781},
  year={2020}
}

@article{harris1954distributional,
  title={Distributional structure},
  author={Harris, Zellig S},
  journal={Word},
  volume={10},
  number={2-3},
  pages={146--162},
  year={1954},
  publisher={Taylor \& Francis}
}

@article{neelakantan2022text,
  title={Text and code embeddings by contrastive pre-training},
  author={Neelakantan, Arvind and Xu, Tao and Puri, Raul and Radford, Alec and Han, Jesse Michael and Tworek, Jerry and Yuan, Qiming and Tezak, Nikolas and Kim, Jong Wook and Hallacy, Chris and others},
  journal={arXiv preprint arXiv:2201.10005},
  year={2022}
}

@article{tang2026kv,
  title={KV-Embedding: Training-free Text Embedding via Internal KV Re-routing in Decoder-only LLMs},
  author={Tang, Yixuan and Yang, Yi},
  journal={arXiv preprint arXiv:2601.01046},
  year={2026}
}

@article{gillin2026bert,
  title={BERT-JEPA: Reorganizing CLS Embeddings for Language-Invariant Semantics},
  author={Gillin, Taj and Lalani, Adam and Zhang, Kenneth and Salles, Marcel Mateos},
  journal={arXiv preprint arXiv:2601.00366},
  year={2026}
}

@article{lei2025making,
  title={Making Large Language Models Efficient Dense Retrievers},
  author={Lei, Yibin and He, Shwai and Li, Ang and Yates, Andrew},
  journal={arXiv preprint arXiv:2512.20612},
  year={2025}
}

@article{xu2025comlq,
  title={ComLQ: Benchmarking Complex Logical Queries in Information Retrieval},
  author={Xu, Ganlin and Yin, Zhitao and Zhang, Linghao and Liang, Jiaqing and Lu, Weijia and Zhang, Xiaodong and Yang, Zhifei and Jiang, Sihang and Yang, Deqing},
  journal={arXiv preprint arXiv:2511.12004},
  year={2025}
}

@inproceedings{janeiro2025mixture,
  title={Mixture of Languages: Improved Multilingual Encoders Through Language Grouping},
  author={Janeiro, Jo{\~a}o Maria and Alastruey, Belen and Massa, Francisco and Elbayad, Maha and Piwowarski, Benjamin and Gallinari, Patrick and Barrault, Loic},
  booktitle={Proceedings of the 2025 Conference on Empirical Methods in Natural Language Processing},
  pages={29695--29710},
  year={2025}
}

@inproceedings{li2025sentence,
  title={Sentence Smith: Controllable Edits for Evaluating Text Embeddings},
  author={Li, Hongji and Michail, Andrianos and Gubelmann, Reto and Clematide, Simon and Opitz, Juri},
  booktitle={Proceedings of the 2025 Conference on Empirical Methods in Natural Language Processing},
  pages={26439--26456},
  year={2025}
}

@article{oda2025one,
  title={One Sentence, Two Embeddings: Contrastive Learning of Explicit and Implicit Semantic Representations},
  author={Oda, Kohei and Chuang, Po-Min and Shirai, Kiyoaki and Kertkeidkachorn, Natthawut},
  journal={arXiv preprint arXiv:2510.09293},
  year={2025}
}

@article{chen2025reasonembed,
  title={Reasonembed: Enhanced text embeddings for reasoning-intensive document retrieval},
  author={Chen, Jianlyu and Lan, Junwei and Li, Chaofan and Lian, Defu and Liu, Zheng},
  journal={arXiv preprint arXiv:2510.08252},
  year={2025}
}

@article{gui2025search,
  title={Search-R3: Unifying Reasoning and Embedding Generation in Large Language Models},
  author={Gui, Yuntao and Cheng, James},
  journal={arXiv preprint arXiv:2510.07048},
  year={2025}
}

@article{sun2025grace,
  title={GRACE: Generative Representation Learning via Contrastive Policy Optimization},
  author={Sun, Jiashuo and Liu, Shixuan and Su, Zhaochen and Zhong, Xianrui and Jiang, Pengcheng and Jin, Bowen and Li, Peiran and Shi, Weijia and Han, Jiawei},
  journal={arXiv preprint arXiv:2510.04506},
  year={2025}
}

@article{ayad2025compressed,
  title={Compressed Concatenation of Small Embedding Models},
  author={Ayad, Mohamed Ayoub Ben and Dinzinger, Michael and Dastidar, Kanishka Ghosh and Mitrovic, Jelena and Granitzer, Michael},
  journal={arXiv preprint arXiv:2510.04626},
  year={2025}
}

@article{zhang2025your,
  title={Your Dense Retriever is Secretly an Expeditious Reasoner},
  author={Zhang, Yichi and Bai, Jun and Cai, Zhixin and Qin, Shuhan and Chen, Zhuofan and Guan, Jinghua and Rong, Wenge},
  journal={arXiv preprint arXiv:2510.21727},
  year={2025}
}

@article{vera2025embeddinggemma,
  title={Embeddinggemma: Powerful and lightweight text representations},
  author={Vera, Henrique Schechter and Dua, Sahil and Zhang, Biao and Salz, Daniel and Mullins, Ryan and Panyam, Sindhu Raghuram and Smoot, Sara and Naim, Iftekhar and Zou, Joe and Chen, Feiyang and others},
  journal={arXiv preprint arXiv:2509.20354},
  year={2025}
}

@article{nacar2025gate,
  title={GATE: General Arabic Text Embedding for Enhanced Semantic Textual Similarity with Matryoshka Representation Learning and Hybrid Loss Training},
  author={Nacar, Omer and Koubaa, Anis and Sibaee, Serry and Al-Habashi, Yasser and Ammar, Adel and Boulila, Wadii},
  journal={arXiv preprint arXiv:2505.24581},
  year={2025}
}

@inproceedings{li2025conan,
  title={Conan-embedding-v2: Training an llm from scratch for text embeddings},
  author={Li, Shiyu and Tang, Yang and Liu, Ruijie and Chen, Shi-Zhe and Chen, Xi},
  booktitle={Proceedings of the 2025 Conference on Empirical Methods in Natural Language Processing},
  pages={15011--15027},
  year={2025}
}

@article{wu2025sitemb,
  title={Sitemb-v1. 5: Improved context-aware dense retrieval for semantic association and long story comprehension},
  author={Wu, Junjie and Li, Jiangnan and Li, Yuqing and Liu, Lemao and Xu, Liyan and Li, Jiwei and Yeung, Dit-Yan and Zhou, Jie and Yu, Mo},
  journal={arXiv preprint arXiv:2508.01959},
  year={2025}
}

@article{lin2025causal2vec,
  title={Causal2Vec: Improving Decoder-only LLMs as Versatile Embedding Models},
  author={Lin, Ailiang and Li, Zhuoyun and Funakoshi, Kotaro and Okumura, Manabu},
  journal={arXiv preprint arXiv:2507.23386},
  year={2025}
}

@article{zhang2025role,
  title={On the role of pretrained language models in general-purpose text embeddings: A survey},
  author={Zhang, Meishan and Zhang, Xin and Zhao, Xinping and Huang, Shouzheng and Hu, Baotian and Zhang, Min},
  journal={arXiv preprint arXiv:2507.20783},
  year={2025}
}

@inproceedings{ranasinghe2025musts,
  title={MUSTS: MUltilingual semantic textual similarity benchmark},
  author={Ranasinghe, Tharindu and Hettiarachchi, Hansi and Orasan, Constantin and Mitkov, Ruslan},
  booktitle={Proceedings of the 63rd Annual Meeting of the Association for Computational Linguistics (Volume 2: Short Papers)},
  pages={331--353},
  year={2025}
}

@inproceedings{chen2025sticking,
  title={Sticking to the Mean: Detecting Sticky Tokens in Text Embedding Models},
  author={Chen, Kexin and Wang, Dongxia and Liu, Yi and Zhang, Haonan and Wang, Wenhai},
  booktitle={Proceedings of the 63rd Annual Meeting of the Association for Computational Linguistics (Volume 1: Long Papers)},
  pages={28660--28681},
  year={2025}
}

@inproceedings{sancheti2025less,
  title={Less Mature is More Adaptable for Sentence-level Language Modeling},
  author={Sancheti, Abhilasha and Dale, David and Kozhevnikov, Artyom and Elbayad, Maha},
  booktitle={Proceedings of the 63rd Annual Meeting of the Association for Computational Linguistics (Volume 1: Long Papers)},
  pages={11680--11695},
  year={2025}
}

@inproceedings{pan2025negative,
  title={Negative matters: Multi-granularity hard-negative synthesis and anchor-token-aware pooling for enhanced text embeddings},
  author={Pan, Tengyu and Duan, Zhichao and Li, Zhenyu and Dong, Bowen and Liu, Ning and Li, Xiuxing and Wang, Jianyong},
  booktitle={Proceedings of the 63rd Annual Meeting of the Association for Computational Linguistics (Volume 1: Long Papers)},
  pages={31102--31118},
  year={2025}
}

@inproceedings{manchanda2025name,
  title={What is in a name? Mitigating Name Bias in Text Embedding Similarity via Anonymization},
  author={Manchanda, Sahil and Shivaswamy, Pannaga},
  booktitle={Findings of the Association for Computational Linguistics: ACL 2025},
  pages={17759--17781},
  year={2025}
}

@article{truong2025learning,
  title={Learning Robust Negation Text Representations},
  author={Truong, Thinh Hung and Verspoor, Karin and Cohn, Trevor and Baldwin, Timothy},
  journal={arXiv preprint arXiv:2507.12782},
  year={2025}
}

@inproceedings{man2025lusifer,
  title={LUSIFER: Language Universal Space Integration for Enhanced Representation in Multilingual Text Embedding Models},
  author={Man, Hieu and Ngo, Nghia Trung and Dac Lai, Viet and Rossi, Ryan A and Dernoncourt, Franck and Huu Nguyen, Thien},
  booktitle={Proceedings of the 48th International ACM SIGIR Conference on Research and Development in Information Retrieval},
  pages={1360--1370},
  year={2025}
}

@article{ponwitayarat2025sea,
  title={SEA-BED: Southeast Asia Embedding Benchmark},
  author={Ponwitayarat, Wuttikorn and Ng, Raymond and Montalan, Jann Railey and Aung, Thura and Ngui, Jian Gang and Susanto, Yosephine and Tjhi, William and Tasawong, Panuthep and Cambria, Erik and Chuangsuwanich, Ekapol and others},
  journal={arXiv preprint arXiv:2508.12243},
  year={2025}
}

@article{lippmann2025zero,
  title={Zero-Shot Contextual Embeddings via Offline Synthetic Corpus Generation},
  author={Lippmann, Philip and Yang, Jie},
  journal={arXiv preprint arXiv:2506.23662},
  year={2025}
}

@article{chung2025maintaining,
  title={Maintaining MTEB: Towards Long Term Usability and Reproducibility of Embedding Benchmarks},
  author={Chung, Isaac and Kerboua, Imene and Kardos, Marton and Solomatin, Roman and Enevoldsen, Kenneth},
  journal={arXiv preprint arXiv:2506.21182},
  year={2025}
}

@article{romero2025beyond,
  title={Beyond instruction-conditioning, MoTE: Mixture of Task Experts for Multi-task Embedding Models},
  author={Romero, Miguel and Ding, Shuoyang and Barret, Corey D and Dinu, Georgiana and Karypis, George},
  journal={arXiv preprint arXiv:2506.17781},
  year={2025}
}

@inproceedings{feng-etal-2025-dont,
    title = "Don{'}t Reinvent the Wheel: Efficient Instruction-Following Text Embedding based on Guided Space Transformation",
    author = "Feng, Yingchaojie  and
      Sun, Yiqun  and
      Sun, Yandong  and
      Zhu, Minfeng  and
      Huang, Qiang  and
      Tung, Anthony Kum Hoe  and
      Chen, Wei",
    booktitle = "Proceedings of the 63rd Annual Meeting of the Association for Computational Linguistics (Volume 1: Long Papers)",
    month = jul,
    year = "2025",
    doi = "10.18653/v1/2025.acl-long.1196",
    pages = "24511--24525",
}

@inproceedings{sun-etal-2025-prism,
    title = "{PRISM}: A Framework for Producing Interpretable Political Bias Embeddings with Political-Aware Cross-Encoder",
    author = "Sun, Yiqun  and
      Huang, Qiang  and
      Tung, Anthony Kum Hoe  and
      Yu, Jun",
    booktitle = "Proceedings of the 63rd Annual Meeting of the Association for Computational Linguistics (Volume 1: Long Papers)",
    month = jul,
    year = "2025",
    pages = "27719--27733",
}

@article{zhang2025gem,
  title={GEM: Empowering LLM for both Embedding Generation and Language Understanding},
  author={Zhang, Caojin and Zhang, Qiang and Li, Ke and Nuthalapati, Sai Vidyaranya and Zhang, Benyu and Liu, Jason and Li, Serena and Zhang, Lizhu and Fan, Xiangjun},
  journal={arXiv preprint arXiv:2506.04344},
  year={2025}
}

@inproceedings{lin-etal-2025-look,
    title = "Look Both Ways and No Sink: Converting {LLM}s into Text Encoders without Training",
    author = "Lin, Ziyong  and
      Wu, Haoyi  and
      Wang, Shu  and
      Tu, Kewei  and
      Zheng, Zilong  and
      Jia, Zixia",
    booktitle = "Proceedings of the 63rd Annual Meeting of the Association for Computational Linguistics (Volume 1: Long Papers)",
    month = jul,
    year = "2025",
    pages = "22839--22853",
}

@article{zhuang2025towards,
  title={Towards Better Instruction Following Retrieval Models},
  author={Zhuang, Yuchen and Trinh, Aaron and Qiang, Rushi and Sun, Haotian and Zhang, Chao and Dai, Hanjun and Dai, Bo},
  journal={arXiv preprint arXiv:2505.21439},
  year={2025}
}

@article{cheng2025contrastive,
  title={Contrastive Prompting Enhances Sentence Embeddings in LLMs through Inference-Time Steering},
  author={Cheng, Zifeng and Wang, Zhonghui and Fu, Yuchen and Jiang, Zhiwei and Yin, Yafeng and Wang, Cong and Gu, Qing},
  journal={arXiv preprint arXiv:2505.12831},
  year={2025}
}

@article{huang2023new,
  title={A new sparse data clustering method based on frequent Items},
  author={Huang, Qiang and Luo, Pingyi and Tung, Anthony KH},
  journal={Proceedings of the ACM on Management of Data},
  volume={1},
  number={1},
  pages={1--28},
  year={2023},
  publisher={ACM New York, NY, USA}
}

@article{li2026weight,
  title={Weight-Informed Self-Explaining Clustering for Mixed-Type Tabular Data},
  author={Li, Lehao and Huang, Qiang and Ang, Yihao and Low, Bryan Kian Hsiang and Tung, Anthony KH and Xiao, Xiaokui},
  journal={arXiv preprint arXiv:2604.05857},
  year={2026}
}

@article{sun2024diversinews,
  title={DiversiNews: Enriching News Consumption with Relevant Yet Diverse News Articles Retrieval},
  author={Sun, Yiqun and Huang, Qiang and Wang, Yanhao and Tung, Anthony KH},
  journal={Proceedings of the VLDB Endowment},
  volume={17},
  number={12},
  pages={4277--4280},
  year={2024},
}

@article{sun2025one,
  title={One Swallow Does Not Make a Summer: Understanding Semantic Structures in Embedding Spaces},
  author={Sun, Yandong and Huang, Qiang and Xu, Ziwei and Sun, Yiqun and Tang, Yixuan and Tung, Anthony KH},
  journal={arXiv preprint arXiv:2512.00852},
  year={2025}
}

@inproceedings{you2026knowledge,
  title={Knowledge Completes the Vision: A Multimodal Entity-aware Retrieval-Augmented Generation Framework for News Image Captioning},
  author={You, Xiaoxing and Huang, Qiang and Li, Lingyu and Zhang, Chi and Liu, Xiaopeng and Zhang, Min and Yu, Jun},
  booktitle={Proceedings of the AAAI Conference on Artificial Intelligence},
  volume={40},
  number={14},
  pages={12108--12116},
  year={2026},
  url={https://ojs.aaai.org/index.php/AAAI/article/view/38200}
}

@article{you2026cut,
  title={{Cut to the Chase: Training-free Multimodal Summarization via Chain-of-Events}},
  author={You, Xiaoxing and Huang, Qiang and Li, Lingyu and Chang, Xiaojun and Yu, Jun},
  journal={arXiv preprint arXiv:2603.06213},
  year={2026}
}

@article{dai2026mg,
  title={{MG$^2$-RAG: Multi-Granularity Graph for Multimodal Retrieval-Augmented Generation}},
  author={Dai, Sijun and Huang, Qiang and You, Xiaoxing and Yu, Jun},
  journal={arXiv preprint arXiv:2604.04969},
  year={2026}
}

@article{sun2026don,
  title={Don't Retrieve, Navigate: Distilling Enterprise Knowledge into Navigable Agent Skills for QA and RAG},
  author={Sun, Yiqun and Wei, Pengfei and Hsieh, Lawrence B},
  journal={arXiv preprint arXiv:2604.14572},
  year={2026}
}

@article{huang2024diversity,
  title={Diversity-Aware $ k $-Maximum Inner Product Search Revisited},
  author={Huang, Qiang and Wang, Yanhao and Sun, Yiqun and Tung, Anthony KH},
  journal={arXiv preprint arXiv:2402.13858},
  year={2024}
}

@inproceedings{zhao2026partially,
  title={Partially shared concept bottleneck models},
  author={Zhao, Delong and Huang, Qiang and Yan, Di and Sun, Yiqun and Yu, Jun},
  booktitle={Proceedings of the AAAI Conference on Artificial Intelligence},
  volume={40},
  number={15},
  pages={13117--13125},
  year={2026}
}

@inproceedings{tang2025uncovering,
  title={Uncovering the bigger picture: Comprehensive event understanding via diverse news retrieval},
  author={Tang, Yixuan and Shi, Yuanyuan and Sun, Yiqun and Tung, Anthony Kum Hoe},
  booktitle={Proceedings of the 2025 Conference on Empirical Methods in Natural Language Processing},
  pages={33927--33945},
  year={2025}
}
